\documentclass[acmsmall]{acmart}
\usepackage[utf8]{inputenc}
\usepackage{array}
\usepackage{longtable}
\usepackage{graphicx}
\usepackage{makecell}
\usepackage{booktabs}
\usepackage{float}
\usepackage{xcolor}
\usepackage{tabularx}
\definecolor{tinymlblue}{RGB}{70,172,200}

\usepackage{xcolor}         
\usepackage[most]{tcolorbox}

\definecolor{tinymlblue}{RGB}{70,172,200}

\newtcolorbox{hardwarebox}[1]{%
  colback      = white,
  colframe     = tinymlblue,
  coltitle     = white,
  colbacktitle = tinymlblue,
  fonttitle    = \bfseries,
  title        = {\textit{#1}}, 
  boxrule      = 0.9pt,
  arc          = 2pt,
  left         = 1.25em,
  right        = 1.25em,
  top          = 0.9em,
  bottom       = 0.9em,
}

\usepackage{xcolor}           
\usepackage{colortbl}         

\definecolor{tinymlblue}{RGB}{70,172,200}  

\DeclareUnicodeCharacter{202F}{\,}

\AtBeginDocument{%
  }

\setcopyright{acmlicensed}
\copyrightyear{2025}
\acmYear{2025}
\acmDOI{XXXXXXX.XXXXXXX}

\acmJournal{JACM}
\acmVolume{37}
\acmNumber{4}
\acmMonth{5}

\begin{document}

\title{From Tiny Machine Learning to Tiny Deep Learning: A Survey}


\author{Shriyank Somvanshi}
\affiliation{%
  \institution{Texas State University}
  \city{San Marcos}
  \country{USA}}
\email{shriyank@txstate.edu}

\author{Md Monzurul Islam}
\affiliation{%
  \institution{Texas State University}
  \city{San Marcos}
  \country{USA}}
\email{monzurul@txstate.edu}

\author{Gaurab Chhetri}
\affiliation{%
  \institution{Texas State University}
  \city{San Marcos}
  \country{USA}}
\email{gaurab@txstate.edu}

\author{Rohit Chakraborty}
\affiliation{%
  \institution{Texas State University}
  \city{San Marcos}
  \country{USA}}
\email{rohitchakraborty@txstate.edu}

\author{Mahmuda Sultana Mimi}
\affiliation{%
  \institution{Texas State University}
  \city{San Marcos}
  \country{USA}}
\email{qnb9@txstate.edu}

\author{Sawgat Ahmed Shuvo}
\affiliation{%
  \institution{Texas State University}
  \city{San Marcos}
  \country{USA}}
\email{sawgat@txstate.edu}

\author{Kazi Sifatul Islam}
\affiliation{%
  \institution{Texas State University}
  \city{San Marcos}
  \country{USA}}
\email{kazi_sifat@txstate.edu}

\author{Syed Aaqib Javed}
\affiliation{%
  \institution{Texas State University}
  \city{San Marcos}
  \country{USA}}
\email{aaqib.ce@txstate.edu}

\author{Sharif Ahmed Rafat}
\affiliation{%
  \institution{Texas State University}
  \city{San Marcos}
  \country{USA}}
\email{sarafat@txstate.edu}

\author{Anandi Dutta, Ph.D.}
\affiliation{%
  \institution{Texas State University}
  \city{San Marcos}
  \country{USA}}
\email{anandi.dutta@txstate.edu}

\author{Subasish Das, Ph.D.}
\affiliation{%
  \institution{Texas State University}
  \city{San Marcos}
  \country{USA}}
\email{subasish@txstate.edu}

\renewcommand{\shortauthors}{Somvanshi et al.}

\begin{abstract}
 The rapid growth of edge devices has driven the demand for deploying artificial intelligence (AI) at the edge, giving rise to Tiny Machine Learning (TinyML) and its evolving counterpart, Tiny Deep Learning (TinyDL). While TinyML initially focused on enabling simple inference tasks on microcontrollers, the emergence of TinyDL marks a paradigm shift toward deploying deep learning models on severely resource-constrained hardware. This survey presents a comprehensive overview of the transition from TinyML to TinyDL, encompassing architectural innovations, hardware platforms, model optimization techniques, and software toolchains. We analyze state-of-the-art methods in quantization, pruning, and neural architecture search (NAS), and examine hardware trends from MCUs to dedicated neural accelerators. Furthermore, we categorize software deployment frameworks, compilers, and AutoML tools enabling practical on-device learning. Applications across domains such as computer vision, audio recognition, healthcare, and industrial monitoring are reviewed to illustrate the real-world impact of TinyDL. Finally, we identify emerging directions including neuromorphic computing, federated TinyDL, edge-native foundation models, and domain-specific co-design approaches. This survey aims to serve as a foundational resource for researchers and practitioners, offering a holistic view of the ecosystem and laying the groundwork for future advancements in edge AI.
\end{abstract}




\received{20 June 2025}

\maketitle

\section{Introduction}
Tiny Machine Learning (TinyML) has emerged as a rapidly growing paradigm that brings machine learning capabilities to severely resource-constrained edge devices. Traditionally, machine learning models demanded significant computational resources, making their deployment on microcontroller units (MCUs) and embedded platforms impractical. However, advances in hardware design, model compression, and embedded inference have allowed real-time intelligence to be embedded on-device, leading to a new class of systems that execute complex analytics at the edge. As the field evolves, a distinct subdomain called Tiny Deep Learning (TinyDL) has gained momentum, focusing specifically on deploying deep learning models, rather than shallow classifiers on low-power, ultra-constrained hardware.


TinyML is typically defined as the deployment of machine learning inference tasks on devices operating under 1\,mW of power, often with only 32 to 512\,kB of Static Random-Access Memory (SRAM) and constrained flash storage. These devices, which usually lack an operating system and hardware accelerators for floating-point operations, are capable of performing real-time analytics while meeting stringent energy and memory budgets~\cite{zaidi2022unlocking, lin2023tiny, han2022tinyml}. TinyDL builds upon this foundation by emphasizing the use of deep neural networks, such as convolutional and transformer-based architectures, under similar constraints. This term, introduced as early as 2017 with just-in-time inference frameworks like TinyDL~\cite{rouhani2017tinydl}, now encompasses a range of state-of-the-art models such as MCUNet, EfficientNet-lite, and DistilBERT variants that deliver strong accuracy with memory footprints below 1\,MB and latency below 20 milliseconds~\cite{le2023efficient}.


The rise of TinyML and TinyDL is primarily driven by limitations inherent in traditional cloud-based machine learning workflows. Cloud inference introduces unacceptable round-trip latencies in time-sensitive applications such as autonomous driving, drones, and wearables~\cite{park2019wireless}. Moreover, transmitting sensor data to the cloud raises substantial privacy concerns in healthcare and industrial Internet of Things (IoT) contexts, where data sovereignty and user trust are paramount~\cite{soro2021tinyml}. Finally, the energy consumption required to constantly stream data to remote servers introduces a prohibitive cost, especially for battery-powered devices~\cite{abadade2023comprehensive}. By shifting inference-and increasingly, lightweight learning-onto the device, TinyDL enables ultra-low-latency responses, reduces dependency on cloud connectivity, and enhances data privacy~\cite{zaidi2022unlocking}.


Initially, TinyML systems relied on shallow models such as linear classifiers, decision trees, or single-layer perceptrons. These models, while lightweight, were unable to match the representational power of deep neural networks and required extensive manual feature engineering, particularly for audio and vision tasks~\cite{ray2022review}. The transition toward TinyDL was made possible by several interrelated advances. First, architectural innovations such as depthwise separable convolutions, inverted residuals, and attention mechanisms made it possible to compress model complexity without sacrificing accuracy~\cite{lin2023tiny}. Second, a suite of optimization techniques including quantization-aware training (QAT), structured pruning, knowledge distillation, and low-rank factorization, dramatically reduced the runtime and memory demands of deep models~\cite{le2023efficient}. Third, the introduction of Neural Architecture Search (NAS) frameworks that co-optimize model topology and deployment constraints-such as MCUNet and TinyNAS-has demonstrated that ImageNet-scale tasks can be executed on MCUs with just 480\, kB of SRAM~\cite{pau2022automated}. Additionally, new developments in on-device and continual learning allow models to adapt in real-time under strict memory and compute constraints, further extending the practicality of TinyDL systems~\cite{delnevo2023evaluation}.

\subsection{Objectives and Scope of This Survey}

This survey aims to provide a comprehensive and timely synthesis of the emerging landscape of TinyDL. While several reviews have previously outlined the evolution of TinyML and its applications up to 2022~\cite{han2022tinyml, abadade2023comprehensive, capogrosso2024machine}, they generally focus on classical machine learning or do not sufficiently distinguish TinyDL as a distinct subfield. Our work addresses this gap by emphasizing deep models and techniques tailored for kilobyte-scale environments. We highlight developments occurring through 2025 and offer an integrated perspective that spans model design, software toolchains, hardware platforms, and deployment strategies. This includes insights from academic research, open-source benchmarks, and industrial deployment case studies. Moreover, we identify critical gaps in current research, such as the lack of support for federated learning, the security of over-the-air updates, and the absence of robust benchmarks for TinyDL systems, and propose a structured agenda for future work.

\subsection{Summary of Contributions}

To guide this discussion, we contribute a unified definition and taxonomy that clearly delineates TinyDL from traditional TinyML, incorporating hardware constraints and algorithmic characteristics. We offer a comprehensive literature synthesis derived from over 200 sources, structured around recent advances in model architectures, NAS methods, toolchains, and application domains. In addition, we propose a benchmarking framework for evaluating TinyDL systems, incorporating metrics such as inference latency, memory usage, model size, and energy efficiency. As a companion to this paper, we also release \texttt{awesome-tinyml}\footnote{\url{https://github.com/gauravfs-14/awesome-tinyml}}, a curated, automatically updated open-source repository of TinyML research papers, tools, frameworks, and tutorials to support community knowledge sharing. Finally, we present a research roadmap that highlights open questions around neuromorphic TinyDL, domain-specific accelerators, compiler–hardware co-design, and privacy-preserving on-device learning. Through these contributions, this survey aims to support both newcomers and experienced researchers in navigating and contributing to the evolving field of TinyDL.

The remainder of this paper is structured as follows: Section~\ref{sec:Background} introduces TinyML and TinyDL concepts; Section~\ref{sec:Hardware} reviews hardware platforms and benchmarks; Section~\ref{sec:Evolution} outlines the evolution from TinyML to TinyDL; Section~\ref{sec:TinyDLArchitectures} presents lightweight deep learning architectures; Section~\ref{sec:SoftwareToolchain} discusses software toolchains and deployment frameworks; Section~\ref{sec:Applications} highlights key applications across domains; Section~\ref{sec:Learning} explores on-device learning methods; Section~\ref{sec:Metrics} covers evaluation metrics and datasets; Section~\ref{sec:Challenges} discusses ongoing research challenges; Section~\ref{sec:Future} suggests future directions; and Section~\ref{sec:conclusion} concludes the paper.


\section{Background and Foundational Concepts}
\label{sec:Background}

\subsection{Tiny Machine Learning}
TinyML has emerged as a transformative field within artificial intelligence, characterized by the deployment and execution of machine learning models on highly resource-constrained embedded devices, particularly MCUs. This paradigm facilitates on-device data processing and inference, thereby pushing intelligence to the very edge of networks \cite{abadade2023comprehensive, capogrosso2024machine}. The fundamental aim of TinyML is to enable sophisticated analytical capabilities directly on hardware platforms that are severely limited in terms of their available resources. In doing so, it supports applications requiring low latency, minimal power consumption, and enhanced data privacy by keeping data local \cite{ray2022review, abadade2023comprehensive, bariah2024large}.

The operational landscape of TinyML is shaped by three critical constraints: memory, power, and compute limitations. Firstly, memory availability is exceptionally scarce. Devices typically include SRAM ranging from a few kilobytes to several hundred kilobytes for runtime operations, and Flash memory often under one megabyte for storing program code and ML models, though in rare cases this may reach up to 2 MB \cite{thingom2023review, capogrosso2024machine, ray2022review, rajapakse2023intelligence, alajlan2022tinyml}. This represents a stark contrast to conventional computing platforms and necessitates the use of highly compact models \cite{thingom2023review}. Secondly, power consumption is a paramount constraint. TinyML devices typically operate on ultra-low power budgets, often in the milliwatt or even microwatt range \cite{abadade2023comprehensive, capogrosso2024machine, alajlan2022tinyml, ray2022review, rajapakse2023intelligence}. This is essential for battery-powered or energy-harvesting systems designed for long-term operation without frequent recharging or maintenance \cite{abadade2023comprehensive, rajapakse2023intelligence}. Thirdly, compute capabilities in these devices are limited. The MCUs generally operate at clock speeds of several tens to a few hundred megahertz, and many lack Floating Point Units (FPUs), which further constrains the deployment of typical ML models unless optimized through quantization techniques \cite{alajlan2022tinyml, capogrosso2024machine, ray2022review}. These hardware limitations necessitate lightweight and efficient models capable of running within constrained environments.

The hardware ecosystem supporting TinyML primarily consists of low-power MCUs integrated with sensors that gather environmental data. Prominent examples include the ARM Cortex-M series, such as the Cortex-M0, M4, and M7, which strike a balance between computational efficiency, power consumption, and cost \cite{ray2022review, han2022tinyml, elhanashi2024advancements}. Other widely used platforms include the STM32 and ESP32 families \cite{alajlan2022tinyml, abadade2023comprehensive}. These MCUs are often paired with application-specific sensors, such as inertial measurement units (IMUs) for motion tracking, microphones for voice command recognition, and low-resolution cameras for vision tasks with constrained compute budgets \cite{han2022tinyml, ray2022review}. This combination of efficient hardware and targeted sensors empowers TinyML to bring intelligence into everyday objects, from wearables to smart infrastructure.

\subsection{Tiny Deep Learning}
TinyDL represents a specialized subfield within the broader TinyML domain, specifically concentrating on the adaptation and deployment of deep learning  models, such as Convolutional Neural Networks (CNNs) and Recurrent Neural Networks (RNNs), onto extremely resource constrained hardware, most notably MCUs \cite{le2023efficient, kornaros2022hardware}. The primary objective of TinyDL is to enable sophisticated tasks like image classification, object detection, and real time gesture recognition on these low power devices \cite{lamaakal2024tinydl, alajlan2022tinyml}.

A key characteristic of TinyDL is the utilization of highly compressed deep models. These models are meticulously optimized to fit within the stringent memory limitations of MCUs, often resulting in model sizes of just a few hundred kilobytes \cite{ahmed2024tinyml, alajlan2024original}. This substantial reduction is achieved through various model compression techniques. Quantization, for instance, involves converting the model's parameters from higher precision floating point numbers to lower precision integers, such as 8 bit integers, thereby shrinking the model size with minimal degradation in accuracy \cite{lamaakal2024tinydl, alajlan2022tinyml, thingom2023review}. Another prevalent technique is pruning, which systematically removes redundant parameters or connections within the neural network to create sparser and more compact models \cite{alajlan2022tinyml, le2023efficient}. Furthermore, TinyDL models are designed for real time inference. This means they can process data and provide outputs almost instantaneously on the device itself, which is crucial for applications requiring immediate responses \cite{lamaakal2024tinydl, alajlan2022tinyml}.

TinyDL differs significantly from conventional Deep Learning. Conventional DL typically relies on powerful computing resources like GPUs and extensive memory (often gigabytes) to train and run large, complex models, with the primary goal of achieving the highest possible accuracy \cite{alajlan2022tinyml, elhanashi2024advancements}. In contrast, TinyDL operates under severe hardware limitations, prioritizing on device efficiency, low power consumption (milliwatts or microwatts), and minimal memory footprint (kilobytes) \cite{alajlan2022tinyml, le2023efficient}. Recent advancements in TinyDL have introduced neural network architectures specifically designed for edge execution, including MobileNet, SqueezeNet, and Tiny-YOLO \cite{khokhlov2020tiny, sanjay2019mobilenet, koonce2021squeezenet}. These models are tailored to execute with fewer floating-point operations and reduced parameter counts, enabling inference in real time even on MCUs without FPUs \cite{capogrosso2024machine, alajlan2022tinyml}. Furthermore, hardware-aware NAS and energy-efficient training paradigms are being actively explored to enhance model deployment on edge platforms \cite{abadade2023comprehensive}. Within the context of TinyML, TinyDL is a specialized area. TinyML encompasses all machine learning techniques, including classical algorithms, that can be deployed on resource-limited devices \cite{immonen2022tiny}. TinyDL, however, specifically addresses the more demanding challenge of implementing and running inherently more complex and resource intensive deep learning models under these same severe constraints \cite{le2023efficient, ahmed2024tinyml}.

\subsection{Workflow in TinyML and TinyDL}
\begin{figure}[htp]
    \centering
    \includegraphics[width=0.98\linewidth]{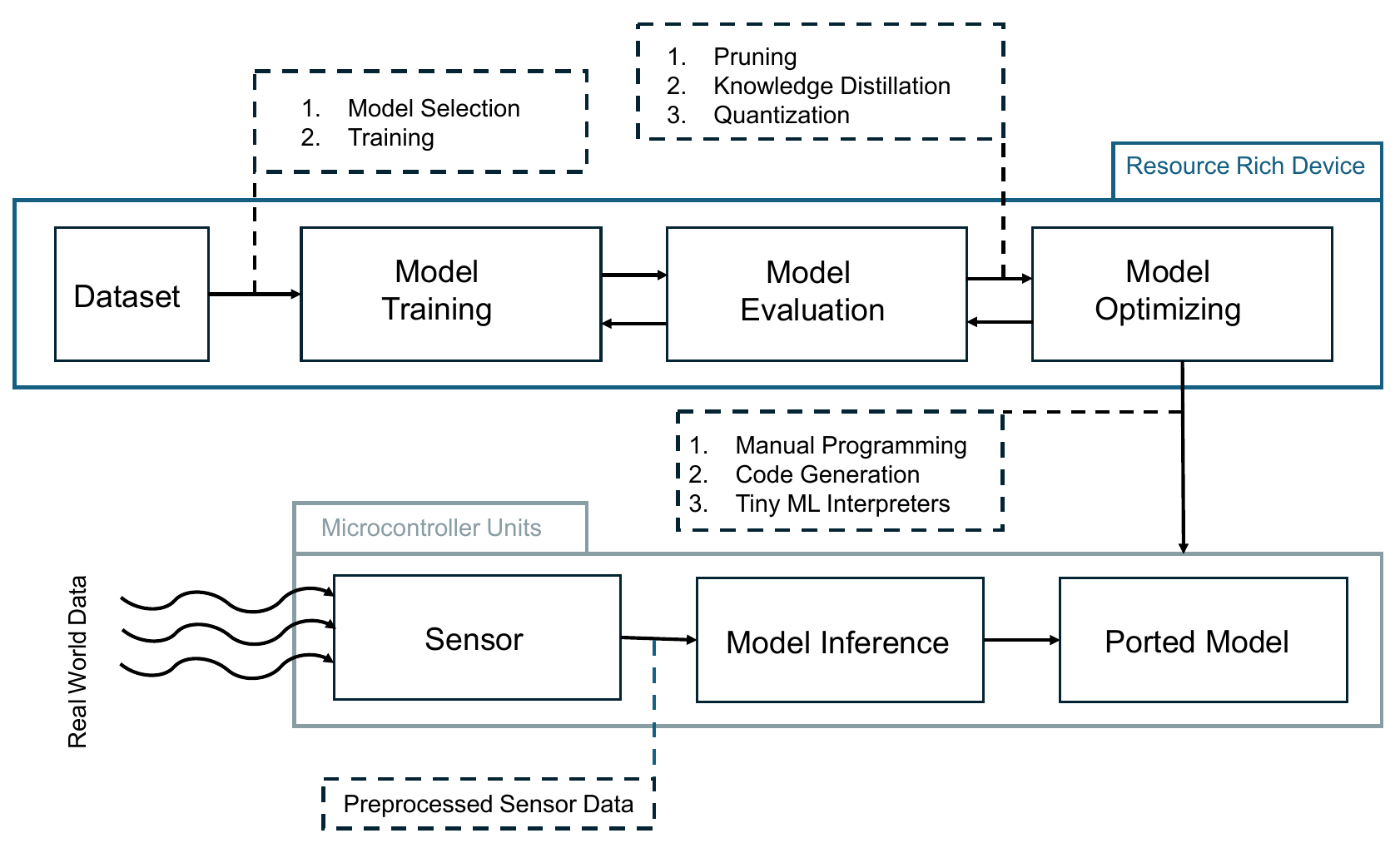}
    \caption{TinyML Pipeline \cite{rajapakse2023intelligence}}
    \label{fig:tinyml_pipline}
\end{figure}

The development of TinyML solutions involves distinct workflows that bridge the creation of machine learning models with their deployment on resource constrained hardware. Three primary approaches are recognized in the literature: the ML oriented workflow, the hardware oriented workflow, and the co design workflow \cite{capogrosso2024machine}.These workflows are illustrated in Figure \ref{fig:tinyml_pipline}, which provides a comparative overview of the design focus, optimization stages, and implementation flow for each approach. The ML oriented workflow is primarily driven by machine learning practitioners. This approach commences with the design, training, and validation of an ML model suited for the specific problem, often initially disregarding the precise limitations of the target hardware to maximize performance and generalization \cite{capogrosso2024machine, rajapakse2023intelligence}. Following this, the model undergoes an optimization phase, where techniques like pruning and quantization are applied to reduce its size and computational demands to meet the hardware constraints \cite{capogrosso2024machine, rajapakse2023intelligence}. The final steps involve deploying the optimized model to the target device and evaluating its real world performance \cite{capogrosso2024machine}.

Conversely, the hardware oriented workflow is led by hardware engineers. The main focus here is on designing new, or enhancing existing, hardware platforms such as MCUs or specialized accelerators, to execute ML algorithms with greater efficiency \cite{capogrosso2024machine}. This workflow involves identifying computational bottlenecks in current hardware when running ML tasks and then architecting specific hardware solutions to mitigate these issues, thereby improving throughput and reducing power consumption for ML workloads \cite{capogrosso2024machine}. The co design workflow represents a more integrated and increasingly pivotal approach. In this paradigm, ML experts and hardware engineers collaborate closely from the project's inception \cite{capogrosso2024machine}. Unlike the sequential nature of the ML oriented or HW oriented workflows, co design involves the intertwined and simultaneous optimization of both the ML model and the hardware architecture \cite{capogrosso2024machine, le2023efficient}. This holistic methodology aims to achieve optimal synergy between software and hardware, potentially unlocking performance and efficiency gains unattainable through isolated optimization efforts, and is considered crucial for advancing the capabilities of TinyML systems \cite{capogrosso2024machine}. Furthermore, a comparative summary of TinyML, TinyDL, and Edge AI paradigms is presented in Table \ref{tab:tinydl_edgeai_tinyml_comparison}, highlighting their design scope, hardware requirements, model complexity, and implementation focus.

\thispagestyle{empty}
\begingroup
\fontsize{9pt}{10pt}\selectfont
\begin{table}[h!]
  \centering
  \caption{Comparison of TinyML, TinyDL, and Edge AI}
  \label{tab:tinydl_edgeai_tinyml_comparison}
  \resizebox{\textwidth}{!}{%
  \begin{tabular}{
    >{\raggedright\arraybackslash}p{0.15\textwidth}
    >{\raggedright\arraybackslash}p{0.283\textwidth}
    >{\raggedright\arraybackslash}p{0.283\textwidth}
    >{\raggedright\arraybackslash}p{0.283\textwidth}
  }
    \toprule
    \rowcolor{tinymlblue}%
    \textcolor{white}{\textbf{Feature}} &
    \textcolor{white}{\textbf{TinyML}} &
    \textcolor{white}{\textbf{TinyDL}} &
    \textcolor{white}{\textbf{Edge AI}} \\
    \midrule
    \textbf{Scope} &
    Subset of Edge AI; ML on extremely resource-constrained devices, typically MCUs~\cite{ray2022review, capogrosso2024machine} &
    Subset of TinyML; specifically deploying deep learning models on these extremely resource-constrained MCUs~\cite{le2023efficient, ahmed2024tinyml} &
    Broadest category; AI processing near the data source, on devices ranging from gateways and edge servers to MCUs~\cite{abadade2023comprehensive, pau2021online} \\
    \textbf{Typical Hardware} &
    MCUs (e.g., ARM Cortex-M series, ESP32), DSPs~\cite{ray2022review, alajlan2022tinyml} &
    Same MCUs, occasionally with minimal accelerators~\cite{lamaakal2024tinydl, elhanashi2024advancements} &
    Edge servers, GPUs (e.g., NVIDIA Jetson), FPGAs, powerful SoCs, capable MCUs/MPUs~\cite{abadade2023comprehensive, ray2022review} \\
    \textbf{Model Complexity} &
    Classical ML and highly compressed DL models (kB to $\lesssim$few MB)~\cite{ray2022review, thingom2023review} &
    Deep networks heavily quantized or pruned to fit kB-scale memory~\cite{ahmed2024tinyml, lamaakal2024tinydl} &
    Ranges from simple ML to complex DL, hardware-dependent~\cite{abadade2023comprehensive}\\
    \textbf{Primary Goals} &
    Enable ML on ultra-low-power, low-cost devices; maximize battery life; ensure data privacy~\cite{ray2022review, abadade2023comprehensive} &
    Bring sophisticated DL to the most constrained devices, pushing efficiency limits~\cite{le2023efficient, alajlan2022tinyml} &
    Reduce latency and bandwidth, improve privacy, enable real-time analytics at the edge~\cite{abadade2023comprehensive, pau2021online}\\
    \textbf{Power Consumption} &
    Typically in the mW–µW range~\cite{abadade2023comprehensive, alajlan2022tinyml} &
    Same as TinyML, with tight per-inference budgets~\cite{le2023efficient, elhanashi2024advancements} &
    Wide range: watts (edge servers) to milliwatts (embedded)~\cite{abadade2023comprehensive}\\
    \textbf{Data Processing} &
    Strictly on-device at MCU level~\cite{ray2022review} &
    Same scope, on-device MCU inference~\cite{lamaakal2024tinydl} &
    Local edge servers, gateways, or capable end devices~\cite{abadade2023comprehensive}\\
    \textbf{Key Characteristic} &
    ML at the “tiniest’’ compute scale, adding intelligence to everyday objects~\cite{ray2022review} &
    Neural networks at kB-scale, requiring extreme optimization~\cite{lamaakal2024tinydl} &
    Decentralized AI that moves computation out of the cloud~\cite{pau2021online}\\
    \textbf{Relationship} &
    Specialized subset of Edge AI, focused on the resource-limited extreme~\cite{abadade2023comprehensive} &
    Specialized subset of TinyML, centered on deep learning algorithms~\cite{ahmed2024tinyml}&
    Superset covering non-cloud AI workloads~\cite{pau2021online}\\
    \bottomrule
  \end{tabular}}
\end{table}
\endgroup

\begin{figure}[htp]
    \centering
    \includegraphics[width=0.98\linewidth]{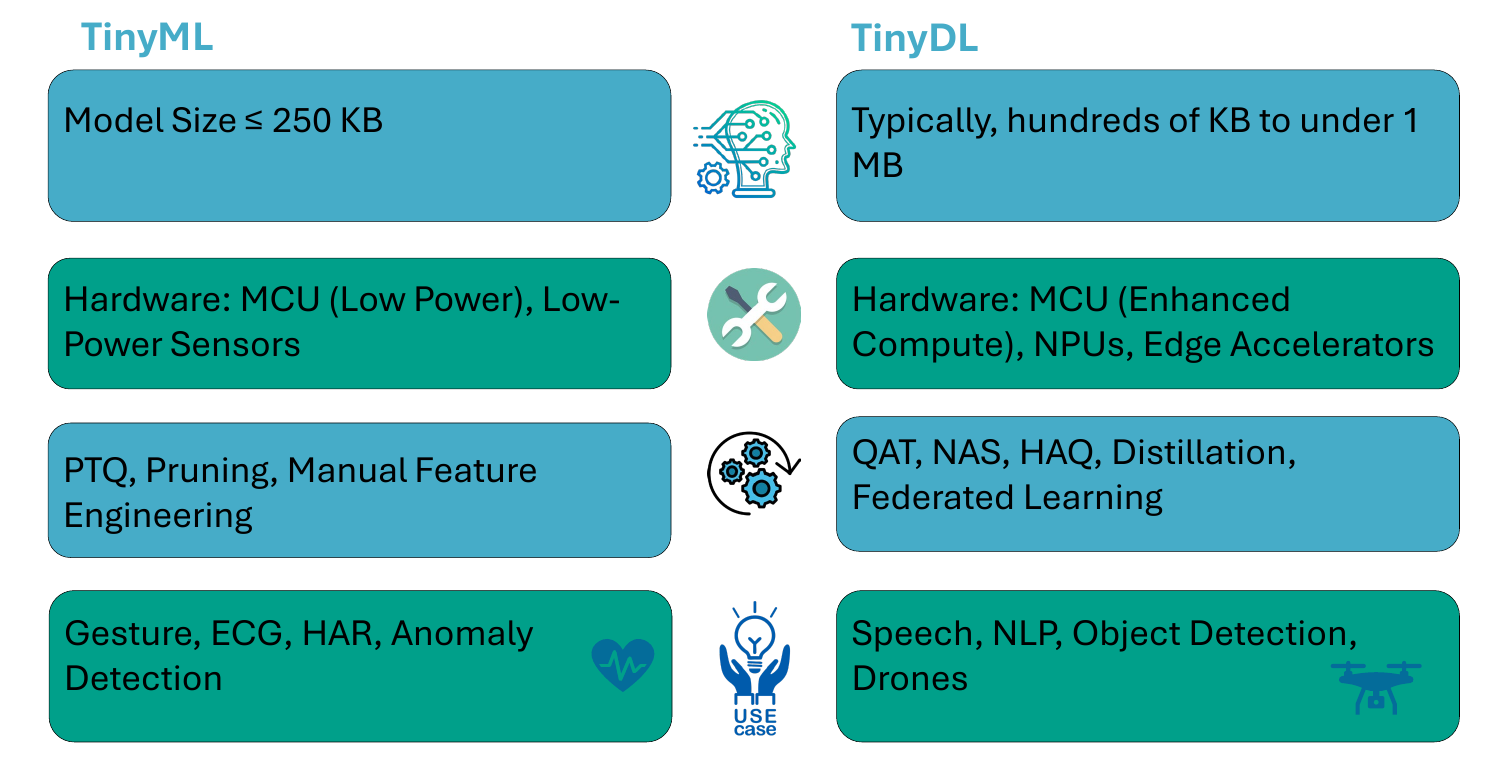}
    \caption{Comparative summary of TinyML and TinyDL across four key aspects: model size, hardware platforms, optimization techniques, and representative use cases}
    \label{fig:TinyMLvsTinyDL}
\end{figure}

To synthesize the transition from TinyML to TinyDL, Figure~\ref{fig:TinyMLvsTinyDL} provides a side-by-side comparison across four key dimensions: model size, hardware platforms, optimization techniques, and representative application domains. While TinyML typically involves deploying classical machine learning models under 250 KB on low-power MCUs, TinyDL enables compressed deep learning models to run on resource-constrained devices through advances in hardware accelerators and model optimization techniques. TinyDL leverages QAT, NAS, hardware-aware quantization (HAQ), and knowledge distillation to maintain high accuracy within stringent memory and energy budgets. As shown, this progression expands the use cases from simple tasks like gesture or electrocardiograms (ECGs) monitoring in TinyML to more complex applications such as speech recognition, vision-based inference, and autonomous systems in TinyDL.

\section{Hardware Platforms for TinyML and TinyDL}
\label{sec:Hardware}

TinyML and TinyDL applications run on highly resource-constrained devices. This section surveys the hardware platforms enabling TinyML, from general MCUs to new specialized AI accelerators, and how these platforms are evaluated. Despite their modest specs, these devices can perform meaningful ML tasks at the edge by balancing performance, power, and accuracy through careful design and benchmarking.

\subsection{MCUs}
MCUs form the backbone of many TinyML deployments, bringing intelligence to the extreme edge. They are single-chip computers designed for low power operation, often running on batteries in remote sensors or wearables \cite{kallimani2024tinyml}. MCUs operate under strict hardware constraints: low clock speeds (typically on the order of tens of MHz) and limited on-chip memory (often only tens to a few hundred kilobytes of RAM). For example, a typical MCU might have approximately 128 KB of RAM and 1 MB of flash storage, versus the gigabytes of memory and storage on a modern smartphone \cite{saha2022machine}. Because they usually run on small batteries, energy efficiency is paramount – TinyML devices must consume mere milliwatts or microwatts to run for long periods \cite{kallimani2024tinyml}.
Despite these limitations, MCUs are capable of running surprisingly complex ML workloads when models are optimized. Continuous improvements in MCU hardware (e.g. more efficient 32-bit ARM Cortex-M processors) have made them powerful and energy-efficient enough to handle small neural networks within tight power budgets \cite{elhanashi2024advancements}. In other words, it is now feasible to deploy machine learning on battery-operated MCU-based sensors in the field. Equally important is the software toolchain that supports MCUs. Frameworks like TensorFlow Lite for MCUs (TFLite Micro) and platforms like Edge Impulse allow developers to compress and deploy trained models on tiny devices. For instance, TFLite Micro can convert a neural network into a form that runs in as little as 16 KB of RAM on an MCU. Techniques such as 8-bit quantization and pruning are used to shrink model size and computation so that inference can execute in real time on limited CPU and memory.\\

\vspace{-10pt}

\begin{hardwarebox}{Key Hardware Constraints for MCU-based TinyML}

\textbf{Clock Speed}: Often 20--200\,MHz clock rates (much lower than GHz-class processors)~\cite{kallimani2024tinyml}, which limits the raw compute throughput of MCUs.

\textbf{Memory}: On the order of kilobytes to a few hundred kilobytes of RAM and usually under a few MB of flash storage~\cite{fauscette_tinyml_2025}. This means models must be very small and efficient.

\textbf{Power/Battery}: Designed for ultra-low power use; many MCU systems run on coin-cell batteries or energy harvesters. Power consumption is in the milliwatt or even microwatt range, so the device can operate for months or years~\cite{fauscette_tinyml_2025}.

\end{hardwarebox}

Even with these constraints, MCUs have demonstrated the ability to run useful ML inference tasks at the edge. By using optimized models, an MCU can perform tasks like keyword spotting (KWS), gesture recognition, anomaly detection, or simple image classification entirely on-device \cite{mandal__2025}. For example, researchers have successfully deployed a voice wake-word detector and simple vision models on tiny boards like the Espressif ESP32 and STMicroelectronics STM32 series MCUs. The ESP32 (a dual-core MCU up to 240 MHz with approximately 520 KB RAM) and various STM32 Cortex-M variants (e.g. an M7 at 216 MHz with a few hundred KB of RAM) are popular choices that, with quantized models, can handle basic deep learning tasks under tight memory and energy constraints. Recent studies and white papers highlight that the combination of improving MCU hardware and clever model optimizations enables these chips to support machine learning workloads that were once thought impossible on such limited devices \cite{kallimani2024tinyml}. MCUs provide a flexible, low-cost platform for TinyML, albeit one that demands extreme efficiency in model design.

\subsection{Specialized AI Hardware}

While many TinyML applications run on general-purpose MCUs, a new class of specialized AI hardware has emerged to push the boundaries of performance and efficiency for tiny and edge deployments. These are application-specific chips and co-processors built specifically to handle neural network inference with minimal energy. By using custom digital logic (and in some cases analog techniques), they act as Neural Compute Engines that drastically accelerate ML tasks compared to a software-only MCU approach. The following subsubsections describe notable examples of AI-focused hardware designed for low-power, on-device deep learning:

\subsubsection{Google Edge Tensor Processing Unit}
The Google Edge Tensor Processing Unit (TPU) is a small application-specific integrated circuit (ASIC) designed by Google to accelerate TensorFlow Lite models at the edge. Each Edge TPU can perform 4 trillion operations per second (4 TOPS) while consuming about 2 W of power (roughly 2 TOPS/W) \cite{syntiant_syntiant_2025}. In practical terms, an Edge TPU can run vision models like MobileNet V2 at nearly 400 frames per second in a power-efficient manner \cite{coral_m2_2025}, far beyond what a typical MCU could achieve. These chips often come as co-processors (e.g., in the Coral EdgeTPU USB sticks or M.2 modules) that pair with a MCU or microprocessor, offloading the heavy math of neural networks. By handling matrix multiplications and convolutions in dedicated hardware, the EdgeTPU enables real-time image and audio inference on the edge device with minimal latency and modest power use.

\subsubsection{Syntiant  Neural Decision Processors Series}
Syntiant’s Neural Decision Processors (NDP) are ultra-low-power neural accelerators aimed at always-on workloads like KWS and sensor analytics. They use a custom deep neural network inference engine that runs models efficiently with parallel multiply-accumulate (MAC) units and an optimized data path for minimal idle cycles \cite{syntiant_syntiant_2025}. For example, the Syntiant NDP120 can continuously listen for voice commands using only a few microwatts. In MLPerf Tiny benchmark tests, the NDP120 was able to perform a KWS inference in about 4.3 ms while consuming only 35~\textmu{}J of energy per inference (at 30 MHz operation) \cite{syntiant_syntiant_2025a}. This is orders of magnitude more energy-efficient than running the same task on a generic MCU. The NDP chips achieve this by being ASICs optimized for neural workloads-they store and process neural network layers on-chip to avoid costly memory accesses, and they integrate small digital signal processor (DSP) cores for preprocessing tasks. Syntiant’s platform demonstrates how specialized silicon can deliver real-time AI within a milliwatt power budget.

\subsubsection{Himax WiseEye WE-I Plus}
The Himax WE-I Plus (HX6537-A) is an example of an AI-enabled MCU/ASIC tailored for vision and sensor inferencing at the edge. It combines a 400 MHz DSP with dedicated hardware accelerators (for tasks like image processing, HOG feature extraction, and JPEG encoding) in an ultra-low-power design \cite{inc_himax_2020}. Uniquely, the WE-I Plus is event-driven: it stays in a near-standby mode until its camera or sensor accelerator detects a trigger (e.g., motion or a person in view), then the DSP wakes to run a neural network inference \cite{inc_himax_2020}. This architecture is highly power-efficient. In fact, when running a person-detection CNN (TinyML vision model), the average power consumption can be under 5 mW-an exceptionally low figure for an image recognition task. By leveraging an ASIC with built-in neural accelerators, the Himax WE-I Plus achieves real-time vision inference (e.g., detecting human presence in a frame) using only a fraction of the energy that a general-purpose MCU would require for the same task \cite{inc_himax_2020}.

These examples illustrate the importance of custom AI silicon for TinyML. Specialized edge AI chips like the Edge TPU, Syntiant NDP, Himax WE-I, as well as others (e.g. Intel’s Movidius Myriad X visual processing unit and various analog neural chips), focus on the common computational patterns of ML algorithms. By implementing neural network operations (matrix multiplies, convolutions, etc.) in hardware, they achieve far higher throughput per watt than a CPU. Innovations such as parallel MAC arrays, on-chip memory for weights/activations, and streamlined dataflows allow these ASICs to perform inference with minimal wasted energy. Many also include features like built-in DSPs or camera interfaces to handle sensor data directly. The result is low-power, real-time inference: tasks like wake-word detection or gesture recognition can run continuously on the edge without exhausting a battery. Table~\ref{tab:tinyml_hw_platforms} summarizes key TinyML hardware platforms, including MCUs and neural accelerators, along with their specifications and typical use cases.

\subsection{Benchmarking and Evaluation}

Standardized benchmarking frameworks play a critical role in enabling fair and consistent evaluation of TinyML hardware platforms. They provide a unified basis for comparing different systems under constraints of speed, power consumption, and inference accuracy. Two widely adopted benchmark suites in this space are MLPerf Tiny and EEMBC MLMark.

MLPerf Tiny, developed by MLCommons in collaboration with EEMBC, is specifically designed for ultra-low-power AI systems. It defines four core tasks representative of common TinyML applications: keyword spotting (KWS), Visual Wake Words (VWW), image classification on low-resolution inputs, and anomaly detection using sensor data \cite{salamone_real-time_2021}. These tasks are performed using compact neural networks, typically under 250 kB in size. MLPerf Tiny evaluates system performance in a single-stream inference mode, mimicking real-time sensor workloads. It reports latency per inference and model accuracy, and also includes an optional energy consumption metric through EEMBC’s EnergyRunner harness \cite{mlcommons_benchmark_2025}. This allows the benchmark to quantify both throughput and power efficiency (e.g., energy per inference), enabling comprehensive comparisons across platforms.

EEMBC MLMark serves a broader purpose by benchmarking general embedded ML inference. It provides a standardized methodology for measuring inference latency and accuracy using fixed models and datasets, ensuring reproducibility across different hardware platforms \cite{mlmark_introducing_2025}. All implementations must use provided test harnesses or disclose optimizations, enforcing a level playing field. While MLMark does not include built-in energy metrics—delegating such measurements to benchmarks like EEMBC’s ULPMark for microcontroller efficiency—it remains valuable for assessing system throughput and model performance \cite{banbury2020benchmarking}.

Together, MLPerf Tiny and MLMark offer complementary strengths. MLPerf Tiny incorporates energy-aware metrics essential for power-constrained TinyML deployments, while MLMark provides broader coverage and a reproducible baseline for embedded ML systems. For example, MLPerf Tiny may reveal that a specialized neural accelerator achieves five times lower latency or ten times greater energy efficiency than a baseline MCU when running the same KWS model \cite{syntiant_syntiant_2025}. Similarly, MLMark can track accuracy and inference improvements as new microcontrollers, neural processing units (NPUs), or software frameworks are introduced \cite{mlmark_introducing_2025}.

These benchmarking tools have become indispensable for researchers and system designers. They enable rigorous, quantitative evaluation of performance, energy usage, and inference quality. As TinyML continues to evolve, these frameworks are expected to adapt by supporting larger models, more diverse sensor modalities, and increasingly fine-grained energy profiling. Such standardized evaluation ensures that TinyML hardware solutions remain responsive to real-world application demands in resource-constrained environments.

\thispagestyle{empty}
\begingroup
\fontsize{9pt}{8pt}\selectfont
\begin{table}[h!]
  \centering
  \caption{Examples of TinyML Hardware Platforms and Their Characteristics}
  \label{tab:tinyml_hw_platforms}
  \resizebox{\textwidth}{!}{%
  \begin{tabular}{  
  >{\raggedright\arraybackslash}p{2.4cm}
  >{\raggedright\arraybackslash}p{2.2cm}
  >{\raggedright\arraybackslash}p{2.8cm}
  >{\raggedright\arraybackslash}p{2.4cm}
  >{\raggedright\arraybackslash}p{3.2cm}
  >{\raggedright\arraybackslash}p{3.0cm}
  }
    \toprule
    \rowcolor{tinymlblue}
    \textcolor{white}{\textbf{Platform}} &
    \textcolor{white}{\textbf{Type}} &
    \textcolor{white}{\textbf{Processor / Clock}} &
    \textcolor{white}{\textbf{Memory (RAM/Flash)}} &
    \textcolor{white}{\textbf{Notable Features}} &
    \textcolor{white}{\textbf{Typical TinyML Use Case}} \\
    \midrule
    Espressif ESP32 & MCU (Wi-Fi SoC) & Dual-core 32-bit MCU @ 240 MHz \newline \cite{kallimani2024tinyml} & 520 KB SRAM, 4 MB flash (external) & Wi-Fi/Bluetooth integrated; low cost & IoT sensors, simple KWS \\
    \addlinespace
    STM32 (e.g., STM32F7) & MCU (Cortex-M7) & 216 MHz ARM Cortex-M7 MCU \newline \cite{kallimani2024tinyml} & $\sim$320 KB RAM, 1 MB flash & DSP instructions, optional FPU & Industrial sensing, audio classification \\
    \addlinespace
    Google EdgeTPU & ASIC Neural Accelerator & Custom ASIC @ $\sim$200 MHz (equiv.) & Uses host memory (external DRAM) & 4 TOPS ($\sim$400 FPS MobileNet) at 2 W \newline PCIe/USB interfaces \newline \cite{coral_m2_2025} & High-speed vision (object detection, etc.) \\
    \addlinespace
    Syntiant NDP120 & Neural co-processor ASIC & Programmable DNN core @ 30–100 MHz & On-chip memory for models & $\sim$35 $\mu$J per inference (KWS); always-on capability \newline \cite{syntiant_syntiant_2025a} & Always-listening AI (wake word, anomaly detection) \\
    \addlinespace
    Himax WE-I Plus & AI MCU/ASIC with DSP & 400 MHz DSP + accelerators \newline \cite{inc_himax_2020} & 2 MB SRAM, 4 MB flash (typical dev board) & Camera interface; $<$5 mW person detection \newline \cite{inc_himax_2020} & Ultra-low-power vision (people counting, etc.) \\
    \bottomrule
  \end{tabular}%
  }
\end{table}
\endgroup


\section{Evolution from TinyML to TinyDL}
\label{sec:Evolution}

\subsection{Limitations of Classical TinyML}
\subsubsection{Poor Generalization for Vision/Audio} Generalization error refers to the difference in performance between a model's training and test datasets. Although this metric is commonly used, it may not fully capture real-world performance, especially when training and test sets come from similar distributions (e.g., same user or device). Research shows that large deep neural networks, despite their capacity to memorize data, often exhibit low generalization error \cite{zhang2017understandingdeeplearningrequires}. In contrast, TinyML models, due to their limited capacity and computational constraints, typically exhibit higher generalization errors. Their reduced ability to learn complex features makes them more prone to poor performance on unseen or out-of-distribution data. For instance, in the VWW task, lightweight convolutional models such as MobileNet and MCUNet have achieved 85–90\% accuracy within 200–250~KB memory budgets \cite{chowdhery2019visual, lin2020mcunet}. In contrast, traditional pipelines using handcrafted features like HOG with classical classifiers (e.g., SVM, decision trees) tend to perform significantly worse-typically in the 70–75\% accuracy range-due to their limited ability to capture spatial hierarchies and generalize to real-world images under constrained memory \cite{lee2020designing, banbury2021mlperf}.

\subsubsection{High Feature Engineering Burden}
Feature engineering involves manually selecting or transforming raw sensor data (e.g., audio, motion, temperature) into meaningful inputs for traditional models like decision trees or SVMs. While essential, this process is time-consuming, requires domain expertise, and often fails to capture complex patterns as effectively as deep learning. In TinyML, these limitations are amplified due to the resource constraints of edge devices, making manual pipelines impractical for real-time, scalable applications. 

\subsection{What has Enabled TinyDL}

\subsubsection{Compression Breakthroughs} Model compression techniques aim to achieve a more efficient representation of one or more layers in a neural network, often with a potential trade-off in quality \cite{menghani2023efficient}. These techniques reduce the model’s size and computational requirements, leading to a 20\% to 30\% decrease in memory usage \cite{abadade2023comprehensive}. There are several general model compression strategies, such as pruning, low-rank factorization, and knowledge distillation. In addition to these, Ray \cite{ray2022review} analyzed specific model compression techniques, including the Tiny Anomaly Compressor \cite{signoretti2021evolving}, Doped Kronecker Product \cite{thakker2020compressing}, and Starfish \cite{hu2020starfish}, particularly in the context of image compression.

Tiny Anomaly Compressor offers a lightweight, model-agnostic compression method that is well-suited for on-device anomaly detection in MCU-based IoT systems, though it faces challenges related to validity and generalizability. Doped Kronecker Product enhances traditional Kronecker Product compression by mitigating accuracy loss in Natural Language Processing (NLP) tasks through co-matrix regularization. Meanwhile, Starfish presents a loss-resilient, AutoML-optimized framework designed for efficient image compression and streaming in resource-constrained IoT environments.

\subsubsection{Quantization and Hardware Advances}

\begin{figure}[htp]
    \centering
    \includegraphics[width=0.98\linewidth]{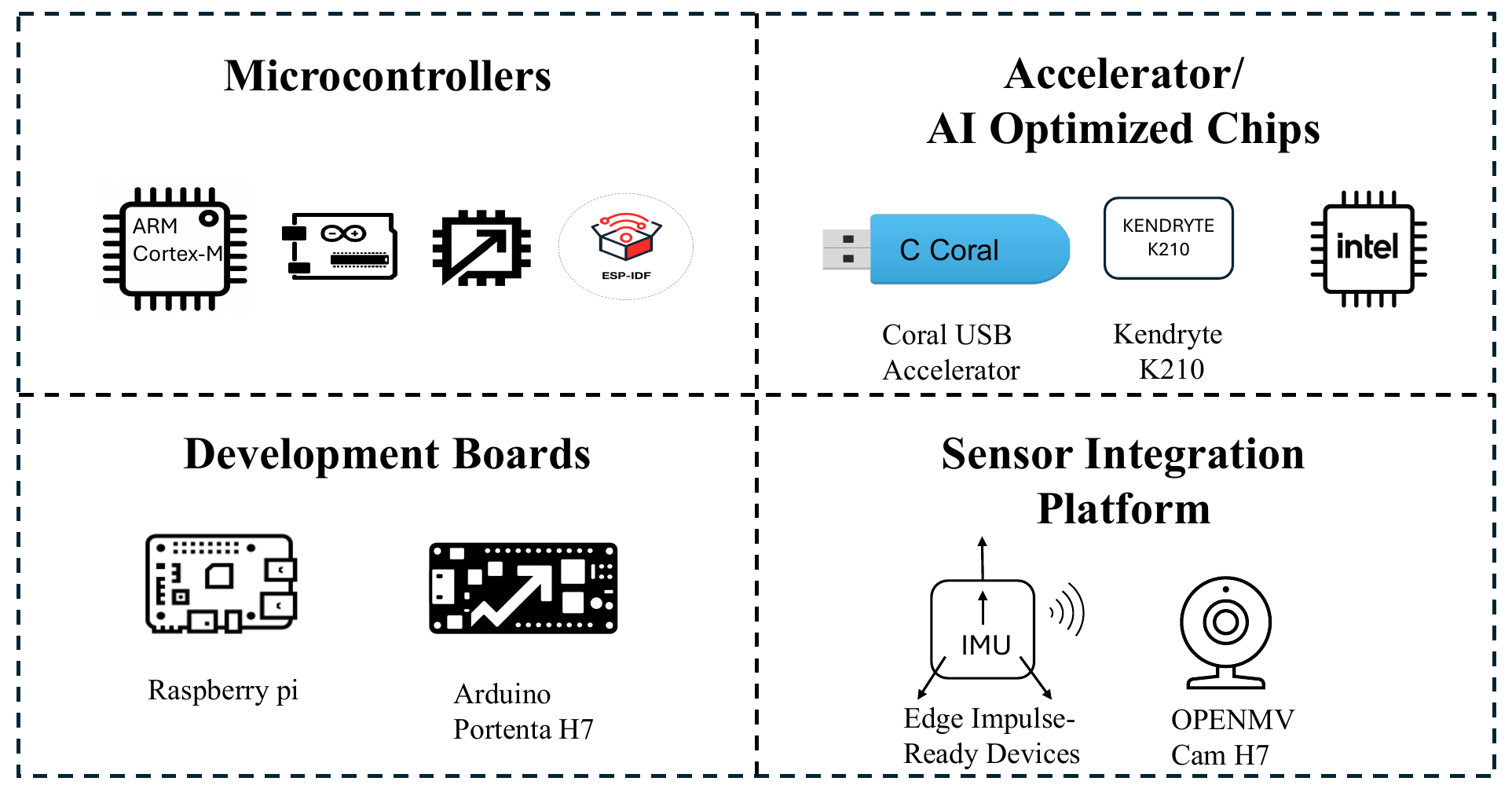}
    \caption{Hardware Ecosystem of TinyDL}
    \label{fig:Hardware}
\end{figure}

Quantization is a cornerstone of TinyDL, enabling significant model compression and computational efficiency by representing weights and activations in reduced-precision formats (typically INT8 or lower). This approach reduces memory footprint and multiply–accumulate (MAC) operations, making deployment feasible on resource-constrained MCUs. Figure~\ref{fig:Hardware} illustrates the TinyDL hardware ecosystem, highlighting core compute units (MCUs, DSPs, NPUs), memory elements (SRAM, Flash, DRAM), and interface components.

Modern quantization techniques go beyond simple format conversion. Hardware-aware approaches and novel numeric representations are increasingly used to optimize for latency and energy. The Cortex-M4, with its Single Instruction Multiple Data/DSP support, remains the most widely adopted MCU in TinyML due to its balance of efficiency and ecosystem maturity. For performance-critical tasks, NPUs accelerate matrix operations with far greater energy efficiency than general-purpose processors.

\begin{hardwarebox}{Key Quantization Approaches Enabling TinyDL}

\textbf{Post-Training Quantization (PTQ):} One-shot float $\rightarrow$ INT8 conversion; \(\sim\)4× size drop with negligible accuracy loss \cite{ray2022review}.

\textbf{Quantization-Aware Training (QAT):} Simulates quant noise during training, enabling 4- or even 2-bit inference while preserving accuracy \cite{chai2021quantization}.

\textbf{Mixed-Precision Schemes (e.g., hardware-aware quantization (HAQ)):} Per-layer bit-width selection (2/4/8 bit) based on latency/energy targets \cite{rusci2020leveraging}.

\textbf{Custom Numeric Formats (e.g., TENT):} Tapered or block-floating formats tuned per layer; up to 31\% energy savings over INT8 baselines \cite{Fatemi2020TENT}.

\end{hardwarebox}

\subsubsection{Lightweight DL Architectures}
Deploying deep neural networks on resource-constrained IoT devices poses significant challenges, particularly in NAS. NAS typically involves three key steps: (i) defining the search space of possible architectures, (ii) applying a search algorithm to find the optimal model, and (iii) using an evaluator to balance accuracy and efficiency for deployment \cite{ray2022review}.

Recent approaches like evolutionary algorithms, differentiable architecture search, progressive search, and parameter sharing have significantly reduced NAS computation costs-from thousands to just a few GPU days-while enabling multi-objective optimization that balances accuracy with efficiency. Techniques such as MNasNet, FBNet, and MONAS further advance this by incorporating latency, power, and computational constraints into the search process, resulting in highly efficient models tailored for deployment on resource-constrained devices \cite{menghani2023efficient}.

\subsection{Key Milestones in TinyDL}
\subsubsection{MobileNet, TinyBERT, MCUNet, SqueezeNet, DistilBERT}
\begin{figure}[htp]
    \centering
    \includegraphics[width=0.98\linewidth]{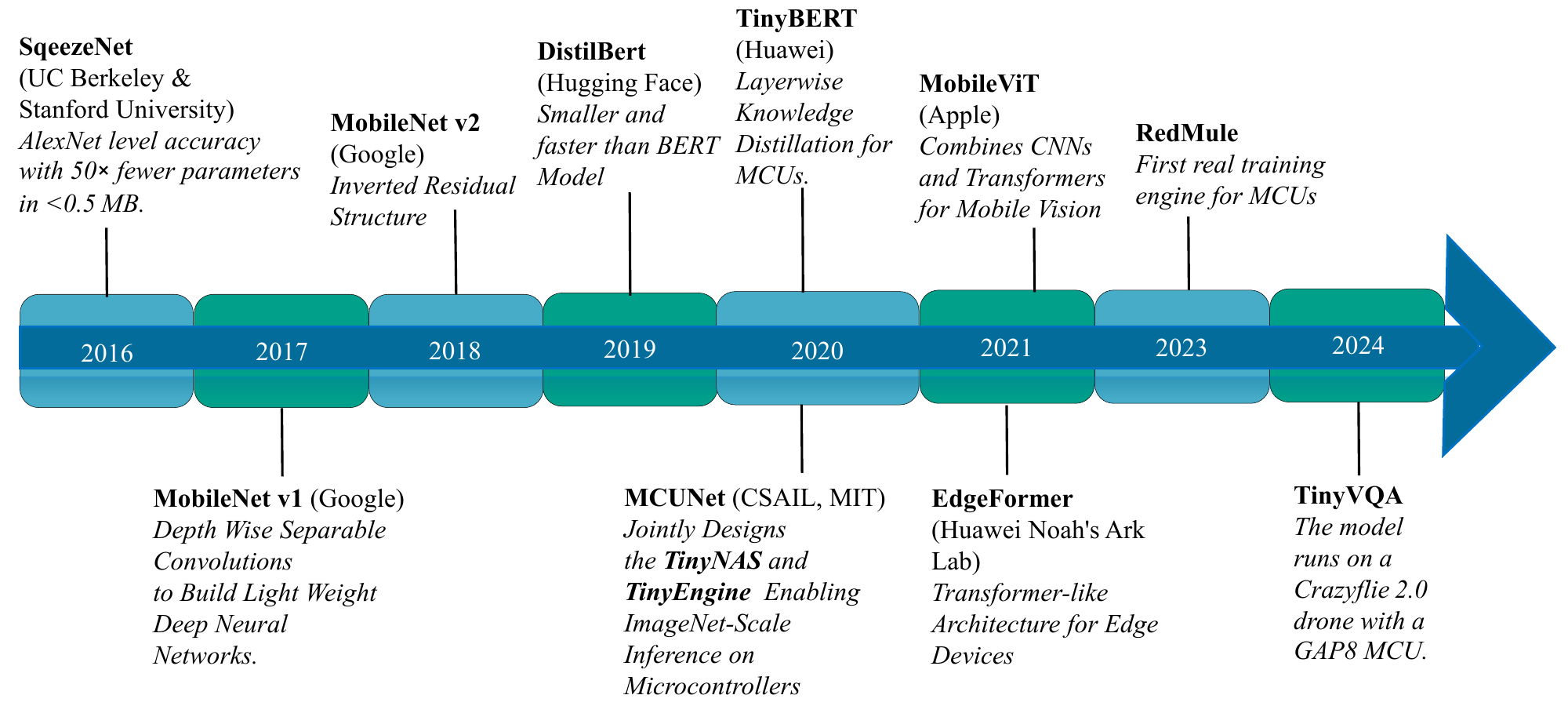}
    \caption{Timeline of Major TinyDL Breakthroughs \cite{iandola2016squeezenetalexnetlevelaccuracy50x},\cite{howard2017mobilenetsefficientconvolutionalneural},\cite{sandler2019mobilenetv2invertedresidualslinear},\cite{sanh2020distilbertdistilledversionbert},\cite{jiao2019tinybert},\cite{lin2020mcunet},\cite{zhang2022parcnetpositionawarecircular},\cite{sandler2019mobilenetv2invertedresidualslinear},\cite{tortorella2023redmulemixedprecisionmatrixmatrixoperation},\cite{rashid2024tinyvqacompactmultimodaldeep}}
    \label{fig:Timeline}
\end{figure}

MobileNet, SqueezeNet, and MCUNet are lightweight models for image tasks, designed to run on devices with limited memory. MobileNet uses depthwise separable convolutions, while SqueezeNet uses fire modules to reduce size. MCUNet goes further by running deep learning on tiny MCUs. For language tasks, TinyBERT and DistilBERT are smaller, faster versions of BERT. TinyBERT uses teacher-student training to keep accuracy high, and DistilBERT keeps 97\% of BERT’s performance with fewer parameters. Figure~\ref{fig:Timeline} summarizes key TinyDL breakthroughs from 2016 to 2024, illustrating the progression from early lightweight CNNs like SqueezeNet and MobileNet to advanced models such as MCUNet, TinyBERT, and RedMule that enable deep learning on microcontrollers.

\section{TinyDL Architectures and Techniques}
\label{sec:TinyDLArchitectures}
The evolution from traditional machine learning to deep learning on resource-constrained devices represents a fundamental paradigm shift in edge computing \cite{heydari2025tiny}. This section examines the architectural innovations and optimization techniques that have enabled the deployment of sophisticated deep learning models on MCU-class devices with severe memory and computational limitations.

\subsection{Lightweight CNNs}

The development of lightweight CNNs specifically designed for resource-constrained environments has been instrumental in enabling deep learning capabilities on edge devices \cite{chand2024empirical}. These architectures employ novel design principles that dramatically reduce computational complexity while maintaining competitive accuracy levels.

\subsubsection{MobileNet Architecture Family}

The MobileNet series represents a cornerstone achievement in efficient CNN design, introducing depthwise separable convolutions that factorize standard convolutions into depthwise and pointwise operations \cite{chand2024empirical}. MobileNetV2 extends this approach with inverted residual blocks and linear bottlenecks, achieving 71.8\% ImageNet top-1 accuracy with only 3.4 MB model size when deployed on STM32H7 MCUs \cite{chand2024empirical}. The architecture's efficiency stems from its fundamental redesign of the convolution operation, reducing parameters by an order of magnitude compared to traditional CNNs while maintaining representational capacity \cite{chand2024empirical}.

\subsubsection{SqueezeNet and Fire Modules}

The SqueezeNet architecture employs fire modules consisting of squeeze layers (1$\times$1 convolutions) followed by expand layers (mixed 1$\times$1 and 3$\times$3 convolutions) to achieve significant parameter reduction \cite{shafiee2017squishednets}. SqueezeNext further optimizes this design with hardware-aware modifications, achieving AlexNet-level accuracy with 112$\times$ fewer parameters \cite{shafiee2017squishednets}. The architecture's aggressive parameter reduction makes it particularly suitable for flash-constrained MCUs, requiring only 1.2 MB storage while maintaining 57.5\% ImageNet accuracy \cite{xu2020electronic}.

\subsubsection{Hardware-Aware Architecture Design}

Recent developments in lightweight CNN design have emphasized hardware-aware optimization through NAS specifically tailored for MCU constraints \cite{banbury2021micronets}. MCUNet demonstrates this approach by achieving 68.7\% ImageNet accuracy with only 0.51 MB model size through joint optimization of network architecture and inference scheduling \cite{lin2020mcunet}. The framework employs a two-stage NAS that first optimizes the search space to fit resource constraints, then specializes the network architecture within the optimized space \cite{lin2020mcunet}.

\subsection{Lightweight Transformers and RNNs}

The adaptation of transformer architectures for TinyML represents a significant advancement in bringing state-of-the-art natural language processing capabilities to edge devices, though it presents unique challenges due to the attention mechanism's quadratic memory complexity \cite{jiao2019tinybert}.

\subsubsection{TinyBERT Knowledge Distillation}

Representing a breakthrough in transformer compression, TinyBERT employs a novel two-stage knowledge distillation framework specifically designed for transformer models \cite{jiao2019tinybert}. The approach performs transformer distillation at both pre-training and task-specific learning stages, enabling effective knowledge transfer from large teacher models to compact student networks \cite{jiao2019tinybert}. TinyBERT-4L achieves 96.7\% performance of BERT-Base on the GLUE benchmark while being 7.5$\times$ smaller and 9.4$\times$ faster on inference, with only 14.5 million parameters \cite{jiao2019tinybert}.

\subsubsection{DistilBERT Compression Strategy}

Utilizing a different distillation approach, DistilBERT reduces the original BERT model by 40\% while retaining 97\% of its language understanding capabilities \cite{jiao2019tinybert}. The model achieves 91.3\% F1 score on SQuAD v1.1 with 66 million parameters, demonstrating the effectiveness of student-teacher training with temperature-scaled softmax distributions \cite{jiao2019tinybert}.

\subsubsection{MCU Deployment Challenges}

Recent research has focused on optimizing transformer deployment specifically for MCU units, addressing unique challenges posed by the multi-head self-attention mechanism \cite{jung2024optimizing}. The primary bottlenecks include high memory footprint of intermediate attention results and frequent data marshaling operations \cite{jung2024optimizing}. Novel approaches such as Fused-Weight Self-Attention (FWSA) and Depth-First Tiling have been developed to mitigate these challenges, achieving up to 6.19$\times$ reduction in memory peak usage while maintaining computational accuracy \cite{jung2024optimizing}.

\bigskip

A comparative summary of popular TinyDL models, including their architectural characteristics, deployment efficiency, and hardware targets, is provided in \textbf{Table~\ref{tab:summary_tiny_models}}. This table highlights the trade-offs between accuracy, model size, and latency across a diverse range of architectures and deployment contexts, offering valuable insights for selecting appropriate models in resource-constrained scenarios.

\thispagestyle{empty}
\begingroup
\fontsize{9pt}{8pt}\selectfont
\begin{table}[htbp]
  \centering
  \caption{Summary of TinyDL Models with Size, Inference Speed, and Task Accuracy}
  \label{tab:summary_tiny_models}
  \resizebox{\textwidth}{!}{%
  \begin{tabular}{
  >{\raggedright\arraybackslash}p{2.4cm}
  >{\raggedright\arraybackslash}p{2.2cm}
  >{\raggedright\arraybackslash}p{1.2cm}
  >{\raggedright\arraybackslash}p{1.2cm}
  >{\raggedright\arraybackslash}p{1.8cm}
  >{\raggedright\arraybackslash}p{2.4cm}
  >{\raggedright\arraybackslash}p{2.4cm}
  }
    \toprule
    \rowcolor{tinymlblue}
    \textcolor{white}{\textbf{Model Name}} &
    \textcolor{white}{\textbf{Architecture}} &
    \textcolor{white}{\textbf{Size (MB)}} &
    \textcolor{white}{\textbf{Latency (ms)}} &
    \textcolor{white}{\textbf{Accuracy (\%)}} &
    \textcolor{white}{\textbf{Target Task}} &
    \textcolor{white}{\textbf{Hardware}} \\
    \midrule
    TinyBERT-4L \cite{jiao2019tinybert} & Transformer & 14.5 & 5.0 & 96.8 (SST-2) & Text Classification & Mobile SoC \\
    \addlinespace
    DistilBERT \cite{jiao2019tinybert} & Transformer & 66.0 & 7.0 & 91.3 (SQuAD) & QA / NLP Tasks & Mobile GPU \\
    \addlinespace
    MobileNetV2-0.35 \cite{chand2024empirical} & CNN & 3.4 & $\sim$32 & 71.8 (ImageNet) & Image Classification & STM32H7 MCU \\
    \addlinespace
    SqueezeNet v1.1 \cite{shafiee2017squishednets,xu2020electronic} & CNN & 4.8 & $\sim$20 & 58.38 (ImageNet) & Object Detection & Kendryte K210 \\
    \addlinespace
    MCUNet-256kB \cite{lin2020mcunet} & CNN + NAS & 0.51 & 12.0 & 70.7 (ImageNet) & Image Classification & STM32F746 \\
    \addlinespace
    EfficientNet-Lite0 \cite{chand2024empirical} & CNN & 4.7 & 45.0 & 75.1 (ImageNet) & Image Classification & EdgeTPU \\
    \addlinespace
    DS-CNN (MLPerf) \cite{banbury2021mlperf} & 1D-CNN & 0.05 & 20.0 & $>$90.0 (Commands) & Wake Word Detection & ARM Cortex-M4 \\
    \addlinespace
    MobileNet (VWW) \cite{banbury2021mlperf} & CNN & 0.32 & 8.0 & 80.0 (VWW) & Visual Wake Words & STM32 MCU \\
    \addlinespace
    Deep AutoEncoder (AD) \cite{banbury2021mlperf} & Autoencoder & 0.27 & 15.0 & 85.0 (AD Bench) & Anomaly Detection & MCU Platform \\
    \addlinespace
    ResNet (IC) \cite{banbury2021mlperf} & CNN & 0.096 & 25.0 & 85.0 (ImageNet) & Image Classification & STM32 MCU \\
    \addlinespace
    Transformer-FWSA \cite{jung2024optimizing} & Transformer & 2.1 & 180.0 & 78.2 (NLP Tasks) & NLP Tasks & STM32F746 \\
    \addlinespace
    SquishedNet \cite{shafiee2017squishednets} & CNN & 0.95 & 156.0 & 77.0 (CIFAR-10) & Image Classification & Nvidia Jetson TX1 \\
    \bottomrule
  \end{tabular}%
  }
\end{table}
\endgroup

\subsection{Model Optimization Techniques}

The deployment of deep learning models on TinyML systems necessitates sophisticated optimization techniques that compress model size and accelerate inference while preserving accuracy \cite{liang2021pruning}.

\subsubsection{Quantization Methodologies}

Quantization represents one of the most effective approaches for model compression in TinyML systems \cite{hawks2021ps}\cite{kuzmin2023pruning}. PTQ converts trained models from floating-point to reduced precision representations, typically INT8, achieving 4$\times$ model size reduction with minimal accuracy degradation \cite{liang2021pruning}. QAT incorporates quantization effects during training, enabling more aggressive precision reduction while maintaining model performance \cite{hawks2021ps}. Comparative analysis shows that quantization generally outperforms pruning across various compression ratios, with benefits becoming more pronounced at moderate compression levels \cite{kuzmin2023pruning}.

\subsubsection{Co-Design of Architecture and Runtime: MCUNet}
\begin{figure}[htp]
    \centering
    \includegraphics[width=0.98\linewidth]{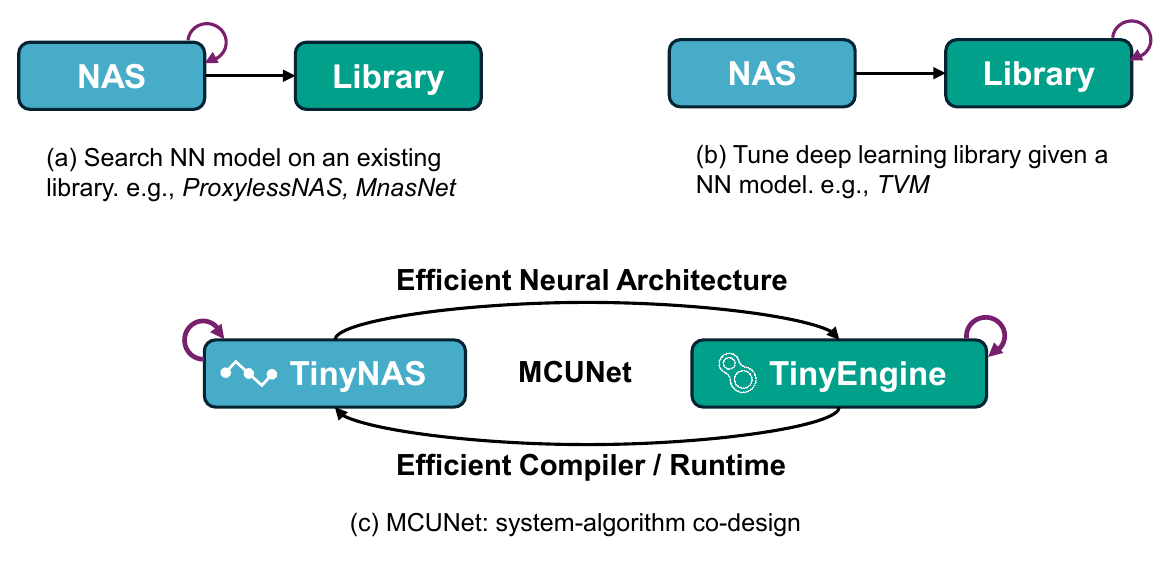}
    \caption{MCUNet co-designs neural architecture (TinyNAS) and inference scheduling (TinyEngine) for MCU efficiency. Unlike (a) architecture search and (b) runtime tuning done separately, (c) MCUNet integrates both for improved accuracy and resource use \cite{lin2020mcunet}}
    \label{fig:mcunet}
\end{figure}

MCUNet exemplifies a system-algorithm co-design approach where both the neural architecture (TinyNAS) and inference engine (TinyEngine) are jointly optimized to meet the extreme memory and compute constraints of MCUs. Unlike traditional pipelines that first fix either the library or the model, MCUNet explores a larger design space by integrating both dimensions. Figure~\ref{fig:mcunet} illustrates this co-optimization flow, highlighting how it surpasses previous approaches limited to one-directional tuning.

\subsubsection{Neural Network Pruning}

Neural network pruning eliminates redundant or less important parameters to reduce model complexity and memory footprint \cite{kumari2024neural}\cite{liang2021pruning}. Magnitude-based pruning removes weights with smallest absolute values, providing a straightforward approach for parameter reduction \cite{kumari2024neural}. Structured pruning targets entire network components such as filters or layers, enabling more significant architectural simplifications suitable for severely resource-constrained environments \cite{liang2021pruning}. Research demonstrates that structured pruning can achieve compression ratios up to 13$\times$ without significant accuracy loss through iterative pruning and retraining cycles \cite{kumari2024neural}.

\subsubsection{Joint Optimization Approaches}

The combination of quantization and pruning techniques has emerged as a powerful strategy for achieving maximum compression efficiency \cite{hawks2021ps}\cite{zhangtraining}. Quantization-aware pruning yields more computationally efficient models than either technique alone, particularly for ultra-low latency applications \cite{hawks2021ps}. Joint optimization frameworks demonstrate superior computational efficiency compared to sequential application of compression techniques, with benefits varying based on target compression ratios and application requirements \cite{zhangtraining}.

\subsubsection{Network Augmentation and Auxiliary Supervision}

\begin{figure}[htp]
    \centering
    \includegraphics[width=0.92\linewidth]{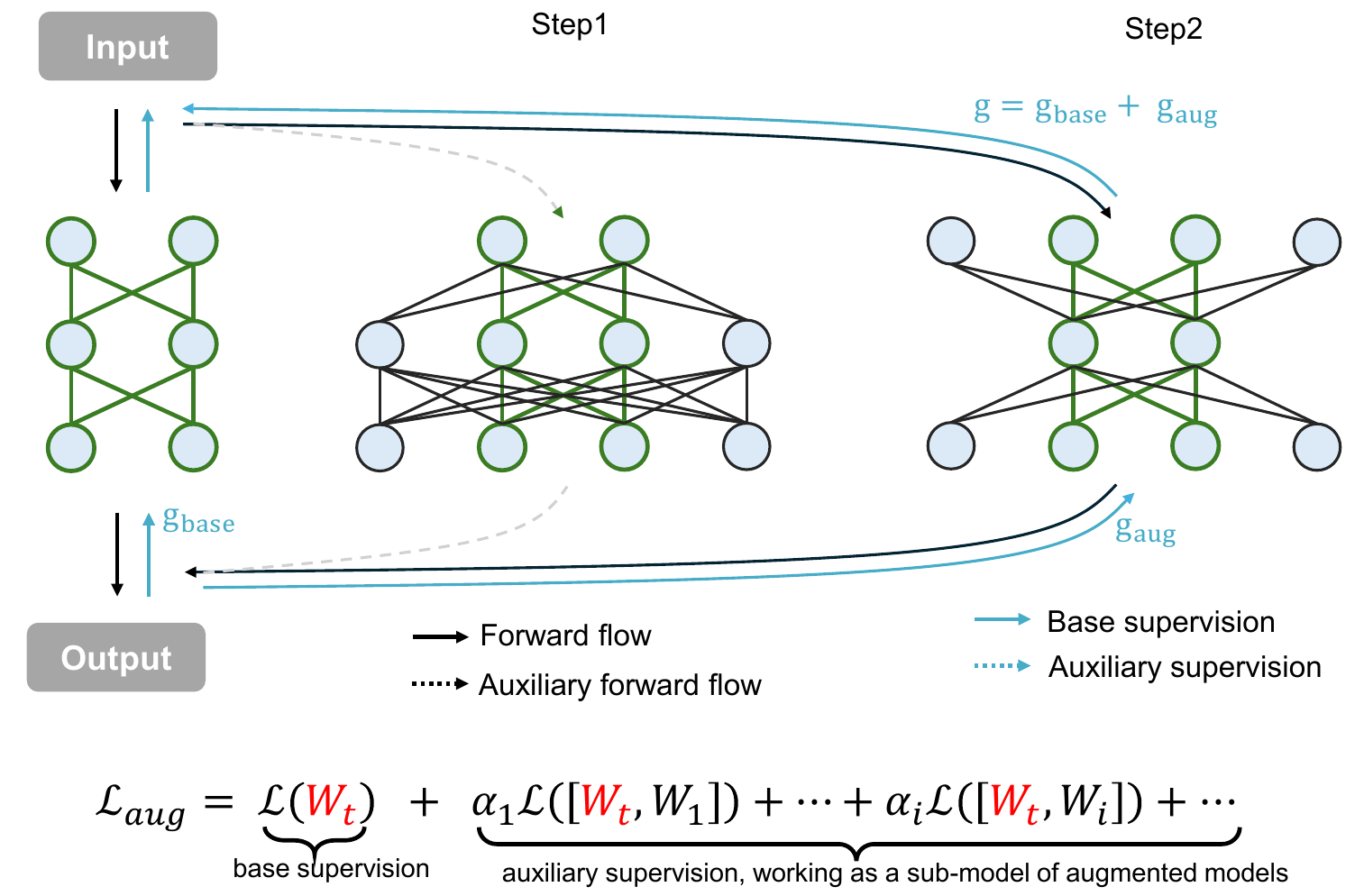}
    \caption{Network augmentation strategy: A tiny model is trained within a larger model to benefit from auxiliary supervision, but only the tiny network is used during inference \cite{cai2021network}.}
    \label{fig:network_aug}
\end{figure}

Network augmentation offers a complementary training-time optimization strategy, particularly relevant for TinyDL. As shown in Figure~\ref{fig:network_aug}, a tiny model is embedded into larger networks that share weights and provide auxiliary supervision. This enables the tiny model to learn stronger representations without increasing its inference-time footprint, making it ideal for resource-constrained deployments.

\section{Software Toolchains and Deployment Frameworks} 
\label{sec:SoftwareToolchain}

The deployment of TinyML and TinyDL models on resource-constrained devices such as MCUs and edge processors requires robust, efficient, and highly optimized software toolchains. These toolchains bridge the gap between trained machine learning models and their real-world deployment on ultra-low-power hardware. This section examines lightweight deployment frameworks, compilation techniques, and end-to-end platforms that enable practical and scalable TinyML and TinyDL implementations.

\subsection{Model Deployment Tools}

This subsection explores a range of lightweight deployment toolchains designed for resource-constrained edge devices, tracing their evolution from early C++ converters to modern platforms with support for quantization, AutoML, and hardware-specific integration. Early frameworks such as uTensor \cite{Tan2018uTensor} demonstrated feasibility by converting TensorFlow models into C++ code but lacked support for quantization or advanced operators. TFLite Micro addressed these limitations by adding support for 8-bit quantization, expanded operator coverage, and community backing, making it a de facto baseline in TinyML deployments \cite{david2021tensorflow}. However, TFLite Micro requires manual tuning and has no built-in GUI. To lower entry barriers, Edge Impulse \cite{Hymel2022EdgeImpulse} introduced a no-code AutoML pipeline with integrated DSP processing and on-device testing. TFLite Model Maker \cite{TFLiteModelMaker} supports fine-tuning and exporting models tailored for deployment on EdgeTPUs and mobile hardware. PyTorch Mobile \cite{PyTorchLite2025}, while not suitable for MCUs, supports deployment of larger TinyDL models (including Transformers) on higher-end mobile SoCs.

More specialized toolchains target performance tuning and low-level integration. CMSIS-NN \cite{Lai2018CMSISNN} provides hand-optimized kernels for ARM Cortex-M architectures and is often paired with TFLite Micro for improved inference latency. MicroTVM \cite{Liu2023MicroTVM}, as an extension of the TVM compilation stack, brings auto-tuning and graph optimization to MCU platforms like STM32 and ESP32. Glow \cite{rotem2018glow}, developed by Meta, offers ahead-of-time graph lowering for hardware accelerators and NPUs. Tools like DeepC convert Keras models into static C code, ideal for systems without dynamic memory support \cite{Keras2C2021}, while MLPACK \cite{Curtin2013MLPACK}, although not TinyML-specific, is a lightweight C++ library adaptable for embedded use. Vendor-specific tools such as X-CUBE-AI \cite{XCubeAI} for STM32 platforms and academic solutions like QKeras with HLS4ML \cite{Duarte2021hls4ml} for FPGA deployment demonstrate the growing ecosystem of domain-targeted solutions. Commercial platforms such as OctoML \cite{ARM2025OctoML} and Nebullvm \cite{Nebullvm2024} further enhance deployment by automating compilation, quantization, and precision tuning across edge platforms. These diverse tools reflect the increasing demand for streamlined, hardware-aware TinyDL deployment pipelines.

\subsection{Compilation and Runtime Support}

To execute models efficiently on tiny hardware, compilation and runtime libraries play a vital role. They optimize memory usage, operator execution, and compatibility with various MCUs, DSPs, and NPUs. CMSIS-NN \cite{Lai2018CMSISNN} was one of the earliest contributions in this space, providing hand-optimized fixed-point kernels for ARM Cortex-M chips. Although it does not support automatic graph compilation, it is frequently paired with TFLite Micro to yield significant gains in latency and power efficiency. To move beyond manually crafted kernels, frameworks such as TVM \cite{1555162} introduced automated compilation workflows that include graph-level optimization, operator fusion, and quantization-aware tuning. Its extension, MicroTVM \cite{Liu2023MicroTVM}, enables deployment on bare-metal devices such as STM32 and ESP32, with support for auto-tuning and memory planning specific to target hardware.

Other frameworks have pursued alternative compilation strategies. Glow \cite{rotem2018glow}, developed by Meta, lowers ML graphs into intermediate representations and compiles them for hardware accelerators, offering static deployment advantages on NPUs. Accelerated Linear Algebra \cite{16He2023ALACompiler}, originally part of TensorFlow, generates backend-specific binaries through operation fusion, and although primarily used in server environments, its techniques influence edge compiler design. ONNX Runtime \cite{17Bai2019ONNX}, when paired with execution providers like TensorRT and OpenVINO, facilitates optimized deployment on edge GPUs and NPUs, though its MCU support remains limited. Emerging tools such as Apache NNC and Tensor Comprehensions \cite{18Vasilache2018TensorComprehensions} explore polyhedral compilation and domain-specific optimization, primarily in research settings. Commercial efforts like OctoML \cite{ARM2025OctoML} offer automated TVM-based tuning services, while Nebullvm \cite{Nebullvm2024} supports quantization, pruning, and cross-platform model optimization. Together, these compilation frameworks ensure that deep learning models, once compressed and optimized, can meet the strict memory and latency constraints of TinyML deployments.

\subsection{End-to-End Platforms}

This section highlights the critical role of compilation and runtime libraries in executing machine learning models efficiently on tiny edge hardware, focusing on tools that enable memory optimization, operator tuning, and hardware-specific acceleration across MCUs, DSPs, and NPUs.

While deployment and compilation tools handle isolated phases of the TinyML lifecycle, end-to-end platforms offer unified environments, with some prioritizing usability and others pushing optimization boundaries. These platforms vary in abstraction, hardware support, and model adaptability, often building on or compensating for the limitations of one another. This section offers an overview of end-to-end TinyML platforms that streamline the entire machine learning lifecycle for edge deployment, while Table \ref{tab:toolchain_comparison} provides a detailed comparison of toolchains and frameworks based on platform support, compression, quantization, AutoML capabilities, and community maturity.

\thispagestyle{empty}
\begingroup
\fontsize{9pt}{8pt}\selectfont
\begin{table}[htbp]
  \centering
  \caption{Comparison of TinyML Toolchains and Frameworks}
  \label{tab:toolchain_comparison}
  \resizebox{\textwidth}{!}{%
  \begin{tabular}{
  >{\raggedright\arraybackslash}p{2.5cm}
  >{\raggedright\arraybackslash}p{2.5cm}
  >{\raggedright\arraybackslash}p{2.5cm}
  >{\raggedright\arraybackslash}p{2.5cm}
  >{\raggedright\arraybackslash}p{2.5cm}
  >{\raggedright\arraybackslash}p{2.5cm}
  }
    \toprule
    \rowcolor{tinymlblue}
    \textcolor{white}{\textbf{Toolchain / Framework}} &
    \textcolor{white}{\textbf{Supported Platforms}} &
    \textcolor{white}{\textbf{Compression Support}} &
    \textcolor{white}{\textbf{Quantization Support}} &
    \textcolor{white}{\textbf{NAS / AutoML Support}} &
    \textcolor{white}{\textbf{Community Maturity / Ecosystem}} \\
    \midrule
    TensorFlow Lite Micro \cite{david2021tensorflow} & MCU (ARM Cortex-M), Arduino, ESP32 & Pruning, weight clustering & 8-bit, int4 (experimental) & Manual only & Very high; large community, strong docs \\
    uTensor \cite{Tan2018uTensor} & ARM Cortex-M, STM32 & Minimal & 8-bit only & No & Low; legacy status, low activity \\
    CMSIS-NN \cite{Lai2018CMSISNN} & ARM Cortex-M family & Manual optimizations & 8-bit fixed-point & No & Medium; well-documented for ARM devs \\
    Edge Impulse Studio \cite{Hymel2022EdgeImpulse} & Web-based IDE (ESP32, STM32, Arduino) & Pruning + DSP fusion & Quant-aware training & Yes (AutoML built-in) & High; growing ecosystem, no-code UI \\
    MicroTVM \cite{Liu2023MicroTVM} & RISC-V, ARM, x86, ESP32 & Compiler-based optimization & INT8/INT4 & Yes (TVM autotuning) & High among compiler researchers \\
    TFLite Model Maker \cite{TFLiteModelMaker} & Android, Raspberry Pi, EdgeTPU & Basic pruning & 8-bit, int16 & Yes (UI-based) & High for mobile/EdgeTPU apps \\
    Qeexo AutoML \cite{qeexo2025userguide} & ARM Cortex-M, IMU boards & Auto-selected models & Yes (auto) & Yes (end-to-end) & Medium; strong for sensor ML \\
    Neuton TinyML \cite{neuton2025ai} & TinyMCU ($<$1 KB RAM) & Highly compressed models & Yes (proprietary) & Yes (Auto-compression) & Emerging; niche but focused \\
    Latent AI \cite{latentai2025} & MCU, CPU, NPU & Pruning, quantization, distillation & Yes (mixed-precision) & Yes (LEIP AutoML) & Medium; commercial/enterprise use \\
    SensiML \cite{sensiml2021overview} & QuickLogic FPGA, MCUs & Auto feature selection & Yes (auto-tuned) & Yes & Growing in sensor-specific domains \\
    OctoML \cite{ARM2025OctoML} & Cloud, Edge (TVM-compatible) & TVM-based optimization & Yes (via TVM) & TVM-integrated AutoTuner & High in TVM community, enterprise users \\
    Arduino IDE / ArduinoML \cite{arduino2025nano33ml} & Arduino boards (AVR, Cortex-M) & None & 8-bit (manual) & No & Large maker community \\
    NXP eIQ Toolkit \cite{nxp2024eiq} & NXP MCUs and NPUs & Quantization, pruning (NXP SDK) & Yes & Limited (GUI only) & Vendor-supported; stable in NXP flow \\
    Microsoft EdgeML \cite{microsoft2025edgeml} & MCU, DSPs (research only) & Model compression (Bonsai, ProtoNN) & Yes & No & Academic; low deployment focus \\
    Google Colab + TFLite \cite{edjeelectronics2025tflite} \cite{edjeelectronics2025github} & Cloud to TFLite-compatible edge & Manual via TFLite converter & Yes & Manual & Broad TF support; no GUI \\
    Sony AI Studio \cite{sony2025ai}\cite{sony2025modelopt} & Sony Spresense board & Pre-configured & Yes (limited) & Model Zoo only & Specialized; Sony-specific \\
    KaaEdge AI \cite{kaaiot2025edgeai} & Edge devices, IoT networks & Model coordination only & Depends on device & Federated AutoML (in development) & System-level orchestration; growing \\
    \bottomrule
  \end{tabular}%
  }
\end{table}
\endgroup


Beyond these widely used frameworks, several specialized platforms provide tailored functionalities that address the needs of different domains. TensorFlow Lite Model Maker \cite{TFLiteModelMaker} offers a Python-based AutoML pipeline with fine-tuning, pruning, and quantization support for deployment on mobile and EdgeTPU devices. For time-series and sensor data, platforms such as Qeexo AutoML \cite{qeexo2025userguide} and SensiML \cite{sensiml2021overview} offer end-to-end workflows that integrate feature extraction, model compression, and deployment to MCUs and FPGAs. These frameworks are especially suited for industrial and automotive applications requiring vibration or IMU-based inference.

To support optimization-focused deployment, platforms such as Latent AI’s LEIP stack \cite{latentai2025} and OctoML \cite{ARM2025OctoML} emphasize model compression workflows across CPUs, NPUs, and MCUs. These frameworks integrate mixed-precision quantization, pruning, and tuning for performance–energy trade-offs. While OctoML builds upon the TVM stack for deployment optimization, Latent AI is designed for hybrid edge use cases requiring cross-platform support.

\begin{hardwarebox}{Popular TinyML Deployment Toolchains Compared}

\textbf{TensorFlow Lite Micro}: de-facto baseline; 8-bit INT quantization, huge community; no GUI \cite{david2021tensorflow}.

\textbf{Edge Impulse Studio}: drag-and-drop AutoML with built-in DSP blocks-ideal for newcomers  \cite{Hymel2022EdgeImpulse}.

\textbf{MicroTVM}: TVM-based compiler autotuning for MCUs (STM32, ESP32) and RISC-V boards \cite{Liu2023MicroTVM}.

\textbf{Neuton TinyML}: sub-kilobyte models that fit where even CMSIS-NN is too heavy  \cite{neuton2025ai}.

\end{hardwarebox}

Other tools address the needs of beginner users or are integrated within hardware-specific ecosystems. Arduino IDE and ArduinoML \cite{arduino2025nano33ml} offer intuitive interfaces for model deployment on AVR and Cortex-M boards but are limited to simpler use cases. The NXP eIQ Toolkit \cite{nxp2024eiq} provides vendor-specific integration for NXP’s MCU and NPU portfolio, offering a graphical interface with built-in support for quantization and pruning.

From the research perspective, Microsoft’s EdgeML \cite{microsoft2025edgeml} introduces novel model architectures such as ProtoNN \cite{10.5555/3305381.3305519} for ultra-low-memory inference. For advanced users, the Google Colab and TFLite workflow \cite{edjeelectronics2025tflite,edjeelectronics2025github} provides maximum scripting flexibility, allowing users to train models in the cloud and convert them for deployment using the TFLite converter. Sony AI Studio \cite{sony2025ai,sony2025modelopt}, while limited to the Spresense board, offers a curated development environment for vision and audio inference. Lastly, KaaEdge AI \cite{kaaiot2025edgeai} addresses large-scale deployment and orchestration needs, supporting federated learning pipelines and distributed edge intelligence.

Together, these platforms span a wide spectrum-from GUI-based environments to optimization-first toolchains-offering developers diverse options depending on their expertise, application complexity, and hardware targets.

\section{Applications of TinyML and TinyDL} 
\label{sec:Applications}

\subsection{Vision}
The field of computer vision has witnessed significant progress, particularly with the advent of deep learning, yet challenges persist in small object detection (SOD) due to factors such as low resolution, limited pixel information, and difficulties in data labeling ~\cite{liang2022deep}. These limitations necessitate specialized approaches, especially when integrating AI into resource-constrained IoT devices, a domain increasingly addressed by TinyML ~\cite{alajlan2022tinyml}. In 2022, research began to address these challenges by exploring various techniques to enhance SOD performance. Methods included augmenting data and implementing super-resolution to improve feature visibility for classifiers ~\cite{liang2022deep}. Furthermore, multi-scale prediction strategies, such as image pyramids and Feature Pyramid Networks, were developed to better adapt to objects of varying sizes and improve detection accuracy ~\cite{liang2022deep}. Additionally, Generative Adversarial Networks were also introduced to boost image resolution and enhance feature representation for small objects~\cite{liang2022deep}. Concurrently, studies focused on improving the energy efficiency and robustness of TinyML computer vision through the use of log-gradient input images, which allow for aggressive quantization and reductions in CNN resources~\cite{lu2022improving}. Software engineering practices for TinyML-based IoT embedded vision were also examined, emphasizing the need for robust and cost-effective solutions for real-world deployments ~\cite{lakshman2022software} . Additionally, challenges and benefits of deploying TinyML on edge devices, including improved privacy and reduced latency, were broadly discussed ~\cite{alajlan2022tinyml}.

Moving into 2023, the focus continued on refining SOD techniques and evaluating TinyML's practical viability. Surveys detailed the significance of SOD in applications like criminal investigation and autonomous driving, categorizing improvement methods such as boosting input resolution and integrating contextual information ~\cite{feng2023deep}. Concurrently, research also began to evaluate the energy feasibility of TinyML for computer vision applications, particularly for tasks like people detection, confirming that TinyML-driven IoT sensors consume less energy compared to traditional machine learning systems ~\cite{denardi2023evaluation}.
More recent advancements, extending into 2024 and 2025, have seen the development of more specialized tools and datasets. The "Wake Vision" dataset was introduced as a large-scale, high-quality benchmark specifically tailored for TinyML computer vision, particularly person detection. This dataset aims to overcome the limitations of smaller, less diverse datasets by using an automated generation pipeline, leading to significant accuracy improvements ~\cite{banbury2025wake}. In parallel, "YoLite+", a lightweight multi-object detection approach, was proposed for traffic scenarios. This method leverages MobileNet and depthwise separable convolution to compress models, achieving faster inference and reduced parameters while maintaining accuracy ~\cite{you2022yolite}. Overall, these developments underscore the ongoing efforts to make TinyML an increasingly practical and powerful tool for diverse computer vision applications in constrained environments.

\subsection{Audio and Natural Language Processing}

The integration of TinyML with NLP has enabled a new class of real-time, energy-efficient AI applications on edge devices, overcoming the limitations of traditional computationally intensive models \cite{barovic2025tinyml,radwan2025tinyml,lamaakal2025comprehensive,pujari2024efficient,jiao2019tinybert}. Efforts in this domain primarily focus on developing lightweight architectures for on-device speech recognition, allowing complex command processing and KWS on MCU platforms such as the Arduino Nano 33 BLE Sense \cite{barovic2025tinyml}. These applications optimize audio signal processing pipelines by incorporating Fast Fourier Transform-based feature extraction while minimizing energy consumption for always-on operation \cite{barovic2025tinyml}. Beyond simple command detection, TinyML-enabled NLP has expanded to more advanced use cases such as semantic sentiment classification using privacy-preserving frameworks like split learning, which reduce computational overhead and enhance data privacy compared to traditional centralized learning approaches \cite{radwan2025tinyml}. Additionally, TinyML is increasingly applied in human-centric contexts, including behavior analysis for smart environments and healthcare, where real-time, privacy-aware data processing is critical despite constrained device resources \cite{lamaakal2025comprehensive}.

These advancements have required adaptation of large transformer-based language models into edge-suitable formats through quantization, pruning, and architectural simplification to meet the demands of low memory, limited processing power, and battery efficiency \cite{rahman2023quantized}. Beyond NLP tasks, TinyML audio models are being used in environmental monitoring systems such as urban noise anomaly detection and wildlife sound recognition. Notably, recent systems like TinyChirp demonstrate efficient bird song classification on low-power wireless acoustic sensors \cite{huang2024tinychirp}. These evolving use cases illustrate the continued maturation of TinyML frameworks and models, showcasing their potential to address complex real-world challenges while operating within severe hardware constraints \cite{kallimani2024tinyml, pujari2024efficient}.

\subsection{Healthcare and Human Behavior Analytics}

TinyML is increasingly transforming healthcare by enabling on-device analytics for various medical applications, prioritizing patient privacy, reducing latency, and enhancing data security \cite{tsoukas2021review}. In particular, TinyML facilitates efficient analysis of ECGs, capturing the heart's electrical signals. A system utilizing reservoir computing on a low-power MCU demonstrated high accuracy with minimal variance \cite{abdennadher2021fixed}. This method lowers complexity and energy consumption while enabling real-time detection of various pathological conditions. Additionally, TinyML-based ECG solutions offer continuous monitoring and instant feedback capabilities for both medical professionals and patients \cite{abadade2023comprehensive}. A practical framework for deploying such real-time health monitoring solutions using MCUs like ESP32 and STM32 was demonstrated by Dutta et al. \cite{dutta2016smart}, emphasizing edge-native inference and system-level integration. Similarly, TinyML allows real-time, low-power respiratory monitoring using a CNN model that detects cancer diseases from acoustic signals with high accuracy. This approach enables remote monitoring, early diagnosis, reduces cloud dependency, and protects privacy, making it ideal for mobile healthcare \cite{genemo2024federated}.

Beyond vital sign monitoring, TinyML is also significantly applied in human behavior analysis within healthcare, notably in emotion detection and Human Activity Recognition. Emotion detection systems utilize wearable devices to analyze physiological signals, including respiratory belts, photoplethysmography, and fingertip temperature \cite{ragot2018emotion, dominguez2020machine}, as well as bioelectrical methods to measure skin conductance, electroencephalography, and heart rate \cite{laureanti2020emotion}. Machine learning models trained on these diverse datasets consistently achieve high accuracies, indicating a strong potential for improving emotional state recognition and optimizing ergonomic conditions \cite{ragot2018emotion,laureanti2020emotion}. For HAR, TinyML enables deployment on resource-limited devices using various algorithms such as CNNs and RNNs like LSTMs. A CNN-based HAR system using mmWave radar, for instance, achieved high accuracy with a minimal model size and fast inference time, addressing privacy and latency concerns associated with camera-based systems \cite{yadav2022tinyradar}. Another notable development is an IoT wristband designed for on-device, privacy-preserving HAR, providing a low-power, low-cost solution that performs real-time activity classification that avoids reliance on cloud infrastructure \cite{saha2023wrist}. Overall, these advancements underscore TinyML's pivotal role in enhancing consumer healthcare data protection through AI-driven and privacy-preserving techniques \cite{johnvictor2024tinyml}, expanding wearable technologies, and supporting intelligent medical devices with high autonomy and energy efficiency \cite{bhamare2024tinyml}.

\subsection{Industrial and Environmental Applications}

TinyML is fundamentally transforming industrial and environmental sectors by enabling efficient, on-device analytics that circumvent the latency, cost, and privacy challenges typically associated with traditional cloud-based machine learning \cite{ooko2024application, ray2022review, dutta2021tinyml, abadade2023comprehensive}. This paradigm shift facilitates real-time industrial anomaly detection by analyzing machine sounds and vibrations directly at the edge, effectively reducing downtime and enhancing operational efficiency \cite{manokaran2022smart}. For instance, specialized TinyML models deployed on ESP-WROOM-32 MCUs are capable of detecting anomalies in thermal images of machinery, transmitting data via Message Queuing Telemetry Transport only when an anomaly is identified and demonstrating high accuracy \cite{oliveira2021edge}. This capability extends to monitoring critical infrastructure, where online learning anomaly detection models like Deep Echo State Network are developed for water distribution systems, adapting to environmental changes directly on MCUs \cite{pau2021online}. Furthermore, deeply quantized anomaly detectors, such as a Block-based Binary Shallow Echo State Network, are proposed for specific industrial use cases like identifying oil leaks in wind turbines, leveraging binarized images and one-bit quantization for efficiency \cite{cardoni2021online}. These applications underscore how TinyML effectively addresses the latency and security vulnerabilities often inherent in cloud-based anomaly detection systems \cite{manokaran2022smart,xenakis2019towards}.

In the broader context of industrial predictive maintenance, TinyML plays a crucial role in preventing costly failures by enabling continuous, on-device monitoring and analysis of equipment data, a domain that continues to see growing research \cite{rajapakse2023intelligence, siang2021anomaly}. Moreover, TinyML also contributes to non-repudiable anomaly detection in extreme industrial settings by integrating blockchain technology, ensuring transparent and immutable records of detected anomalies \cite{antonini2022tinyml}. This integration of TinyML with embedded systems and federated learning in Industrial IoT further aims to decrease latency, increase productivity, and enhance data security in complex manufacturing environments \cite{abubakar2025iiot, casiroli2023tiny}. Additionally, TinyML-enabled smart objects and their associated challenges are also widely discussed as a new paradigm for efficient, privacy-preserving, and cost-effective solutions across various IoT applications \cite{sanchez2020tinyml}.

Beyond industrial applications, TinyML significantly contributes to solving environmental problems, particularly in smart agriculture and wildlife monitoring \cite{bamoumen2022tinyml}. With the global population projected to grow significantly \cite{UN2017}, smart agriculture leverages IoT, drones, and machine learning to enable real-time monitoring of crop and soil health, disease detection, and growth tracking \cite{mitra2022everything, condran2022machine}. TinyML is gaining traction in SA by facilitating machine learning tasks directly on low-power sensor devices, offering reduced latency, enhanced privacy, and lower energy consumption, which is especially valuable in underserved areas with limited connectivity \cite{kalyani2021systematic, abadade2023comprehensive, siang2021anomaly, quy2022iot}. An example is the Nuru app, developed under the PlantVillage project, which utilizes TensorFlow Lite to detect plant diseases offline, aiding farmers in remote regions \cite{gondchawar2016iot, sushanth2018iot}. Additionally, in wildlife conservation, TinyML supports on-animal and bioacoustic sensors for species tracking, addressing challenges in handling latency and data volume in vast habitats \cite{tuia2022perspectives}. This includes real-time tracking of endangered sea turtles using SmallSats \cite{curnick2022smallsats} and TinyML-enabled collars to reduce elephant losses from poaching \cite{hing2024edge, reddy2024edge}. These ongoing advancements and broad survey efforts emphasize TinyML's critical role in advancing industrial efficiency and environmental sustainability by pushing AI capabilities closer to the data source \cite{capogrosso2024machine, lorenzo2024trees}.

\section{On-Device Learning and Reformability} 
\label{sec:Learning}

The evolution of TinyML has opened new frontiers for executing artificial intelligence directly on resource-constrained endpoint devices \cite{dutta2021tinyml, ray2022review}. Indeed, a significant paradigm shift within this domain is the move from static, pre-trained models toward dynamic systems capable of learning and adapting post-deployment \cite{abadade2023comprehensive}. This progression towards on-device learning and the concept of reformability is critical for the long-term viability and effectiveness of TinyML applications, particularly as they become integrated into dynamic, real-world environments \cite{kallimani2024tinyml, rajapakse2023intelligence}. Such capabilities not only enhance model performance over time but also offer substantial benefits for user privacy and data security by keeping sensitive information localized on the device \cite{pujari2023enhancing, lamaakal2024tinydl}.

\subsection{Continual and Few-Shot Learning on MCUs}
Traditional TinyML workflows involve training a machine learning model offline on powerful servers and then deploying a highly optimized, static version for inference on a MCU \cite{dutta2021tinyml}. However, this approach is limited, as the performance of a static model can degrade when the data distribution it encounters in the real world changes over time, a phenomenon known as concept drift \cite{abadade2023comprehensive}. To overcome this limitation, research has increasingly focused on enabling learning capabilities directly on the MCU. A promising solution is the implementation of online or continual learning, where the model can be updated incrementally as new data becomes available. This allows the device to adapt to changing environmental conditions without requiring a complete redeployment of the model \cite{ren2021tinyol}. For instance, the TinyOL framework was proposed to facilitate online learning on MCUs, allowing them to learn from a continuous stream of data \cite{abadade2023comprehensive}. This approach is instrumental for applications that require long-term autonomy and robustness. Furthermore, advancements in few-shot learning are enabling MCUs to train effectively on a very small number of examples. This is particularly relevant for customization and personalization, where a device might need to learn new keywords or commands specific to a user. Research in few-shot KWS has demonstrated that models deployed on MCUs can be trained to recognize new words with only a handful of training samples, making the end-user experience significantly more flexible and interactive \cite{mazumder2021few}. Such on-device training capabilities are often facilitated through methods like federated learning, which allows for collaborative model training across multiple decentralized devices while keeping the raw data localized \cite{kopparapu2021tinyfedtl}.

\subsection{Reformable TinyML}
\begin{figure}[htp]
    \centering
    \includegraphics[width=0.98\linewidth]{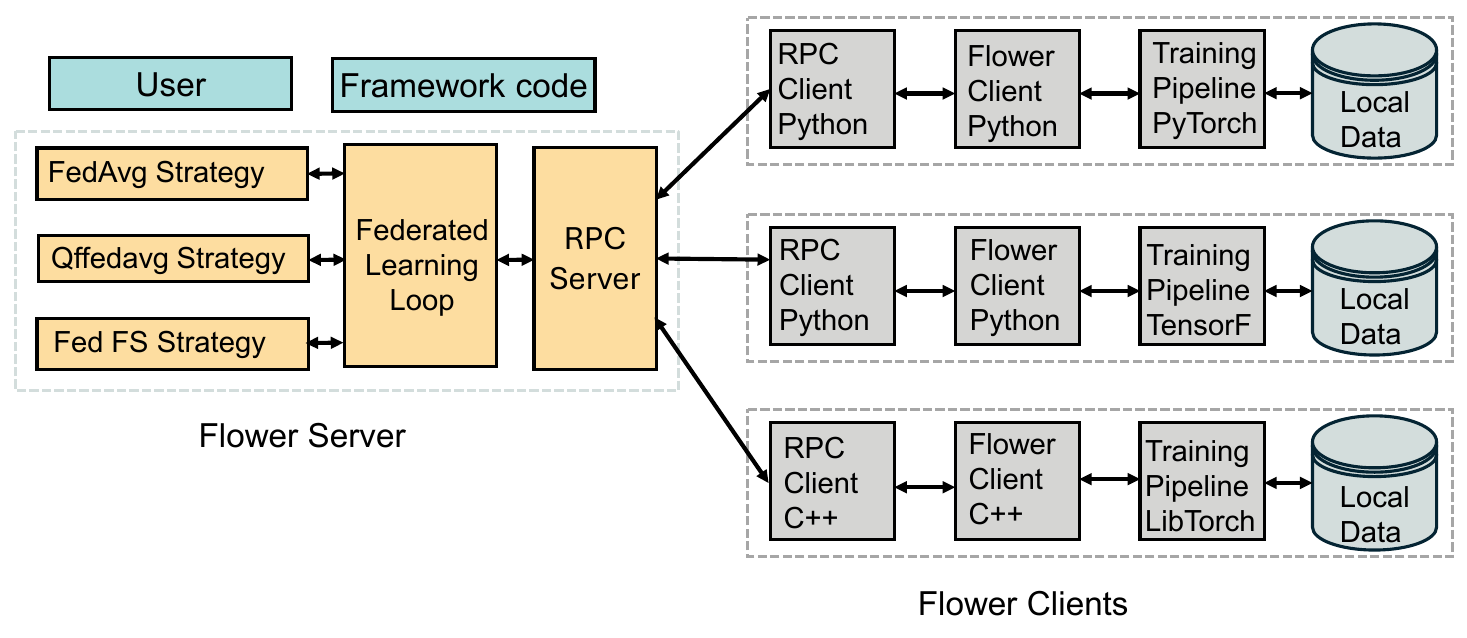}
    \caption{Architecture for Privacy-Preserving Federated TinyML \cite{rajapakse2023intelligence}}
    \label{fig:reformabiltiy}
\end{figure}

Building upon the concept of on-device learning is the emerging paradigm of Reformable TinyML. This refers to a holistic framework where TinyML systems are engineered with the inherent capability to be modified, updated, or reformed after their initial deployment \cite{rajapakse2023intelligence}. The primary goal is to create resilient and sustainable intelligent systems that can self-diagnose performance degradation and trigger an adaptation process to maintain accuracy and efficiency over their operational lifetime \cite{kallimani2024tinyml}. The reformable pipeline extends beyond simple on-device updates. It encompasses a structured workflow that includes monitoring for data drift, evaluating the current model's efficacy, and invoking a reformation strategy when necessary. This strategy could involve on-device fine-tuning, fetching a model patch from an edge server, or participating in a federated learning task to receive an updated global model \cite{rajapakse2023intelligence}. Consequently, reformability ensures that the intelligence at the extreme edge does not become obsolete, thereby enhancing the overall value and reliability of the IoT ecosystem. The architecture of such a pipeline is conceptually illustrated in Figure \ref{fig:reformabiltiy}.

\subsection{Privacy and Security Benefits}
A foundational advantage of processing data at the edge is the inherent enhancement of privacy and security \cite{dutta2021tinyml}. By performing ML inference directly on the MCU, sensitive user data, such as audio from a microphone or images from a camera, does not need to be transmitted to the cloud or a remote server \cite{ray2022review}. This localization of data significantly reduces the risks associated with data breaches during transmission and unauthorized access on third-party servers, a crucial consideration for human-centered applications \cite{lamaakal2024tinydl}. The push for on-device learning and reformability further strengthens these privacy guarantees. In a federated learning context, for example, the raw data used for training never leaves the user's device. Instead, only anonymized model updates or gradients are shared to contribute to a global model, ensuring collaborative learning without compromising individual privacy \cite{grau2021device}. Moreover, recent work has focused on integrating explicit privacy-preserving mechanisms into the TinyML framework. Techniques such as local differential privacy can be implemented to add statistical noise to data or model updates before they are shared, providing mathematical guarantees of privacy \cite{pujari2024efficient}. In conjunction with secure hardware features like memory protection units (MPUs) or secure enclaves to encrypt model parameters, these methods provide a robust, multi-layered approach to securing intelligent edge devices \cite{pujari2024efficient}.

\section{Evaluation Metrics and Benchmarks} 
\label{sec:Metrics}

Evaluating TinyML models requires a holistic framework that addresses the unique constraints of ultra-resource-constrained environments while preserving sufficient task performance for real-world deployment. Unlike conventional machine learning systems evaluated primarily based on accuracy or F1-score, TinyML models must be evaluated using a blend of metrics that reflect efficiency, scalability, deployment feasibility, and trade-offs between model performance and resource utilization. These metrics span a wide spectrum, including computational efficiency (inference latency, throughput), memory and storage footprint (model size, RAM usage), energy consumption (power and battery profile), and task-specific accuracy under quantized or pruned conditions. Moreover, consistent comparison across approaches is facilitated by a small set of curated benchmarks and datasets tailored for the TinyML domain, such as KWS, low-resolution vision tasks, and sensor signal classification.

\subsection{Efficiency and Compression Metrics}
In TinyML, efficiency is a first-class citizen. Edge devices such as ARM Cortex-M MCUs or RISC-V-based embedded systems typically offer only kilobytes of RAM and flash storage, no floating-point hardware, and must run on energy-constrained or even battery-less systems. Consequently, performance metrics in TinyML revolve around the ability to deploy models that fit within these extreme limitations without sacrificing critical functionality. The most frequently used efficiency metrics include:

\subsubsection{Model Size (KB)} Model size, measured in kilobytes, represents the total footprint of the model when stored on the device, including weights, biases, and optionally additional metadata. Given that popular MCUs (e.g., STM32F746) have only 512~KB to 1~MB of flash memory, model size is often capped at 250--500~KB to allow room for the operating system and runtime libraries. Compression techniques such as weight pruning, Huffman encoding, and PTQ are employed to reduce model size without significant degradation in performance. For instance, Han et al.~\cite{han2015deep} demonstrate that deep neural networks can be compressed by 9--13$\times$ using a combination of pruning and quantization, with minimal accuracy drop. MCUNet~\cite{lin2020mcunet}, an architecture designed specifically for MCUs, achieves ResNet-level accuracy on ImageNet with models as small as 300~KB.

\subsubsection{Inference Latency (ms)} Inference latency is the wall-clock time taken to perform one forward pass of the model on the target hardware, typically measured in milliseconds. Latency affects the responsiveness of real-time systems like gesture recognition or wake-word detection. For example, Google's KWS model for TensorFlow Lite Micro executes in less than 20~ms on a Cortex-M4 at 80~MHz~\cite{warden2019tinyml}. Latency is influenced by model depth, width, and data precision (e.g., 8-bit vs. floating-point), and can be reduced using techniques like operator fusion, loop unrolling, and hardware-specific optimizations such as CMSIS-NN~\cite{Lai2018CMSISNN}.

\subsubsection{Memory Usage (KB)} Memory usage includes both the \textit{static memory footprint} (the size of the model parameters) and \textit{dynamic memory} (temporary activations, intermediate buffers). MCUs typically have only tens to hundreds of kilobytes of SRAM; for example, the STM32L475 has just 128~KB of SRAM, which limits the size of intermediate tensors and imposes constraints on model depth and width~\cite{warden2019tinyml}. Memory profiling tools in frameworks such as TensorFlow Lite Micro~\cite{david2021tensorflow} and TVM~\cite{chen2018tvm} provide developers with visibility into memory allocation and enable optimization of memory usage across inference stages. Techniques such as activation recomputation (also known as gradient checkpointing)~\cite{gruslys2016memory} and in-place memory reuse~\cite{cho2017mec} are particularly effective in minimizing peak RAM usage. Additionally, CMSIS-NN~\cite{Lai2018CMSISNN} and other hand-optimized libraries help reduce memory overhead by using fixed-point arithmetic and optimizing buffer allocation for low-latency execution on Arm Cortex-M processors.

\subsubsection{Operations Count (MACs / FLOPs)} Operations count refers to the total multiply-accumulate operations (MACs) or floating-point operations (FLOPs) required for a single inference. It is often used as a proxy for computational workload and correlates strongly with both latency and energy consumption on MCU-class hardware. MLPerf Tiny includes MACs as a primary metric to standardize comparisons across architectures, enabling fair trade-off analysis across accuracy, latency, and energy \cite{banbury2021mlperf}.

\subsubsection{Energy Consumption (mJ)} For battery-powered or energy-harvesting devices, energy efficiency is paramount. Energy per inference, measured in millijoules (mJ), is a product of inference latency and average power consumption. Banbury et al.~\cite{banbury2021mlperf} show that their benchmark models consume between 0.1~mJ and 10~mJ per inference. Specialized accelerators like the GAP8 or Ambiq Apollo chips allow sub-mJ inference for tasks like image classification or speech recognition.

\subsection{Accuracy Trade-offs and Constraints}

Accuracy remains an important performance metric, particularly when TinyML is used in critical applications such as medical diagnostics or industrial anomaly detection. However, due to severe memory and compute limitations, models are often subject to trade-offs between efficiency and accuracy. One of the most widely studied compromises is the balance between quantization and model performance.

\subsubsection{Quantization vs. Accuracy} Quantization reduces the precision of weights and activations, typically from 32-bit floating point (FP32) to 8-bit integers (INT8), and sometimes lower. This can lead to massive reductions in model size, latency, and energy usage, but may affect model accuracy. Magnitude-based structured pruning of MobileNetV2 (50\% weights removed) incurs $<1\%$ top-1 drop on ImageNet while shrinking model size by 1.9 times~\cite{liang2021pruning}. Jacob et al.~\cite{jacob2018quantization} found that PTQ can reduce MobileNetV1 accuracy on ImageNet by up to 3--4\%, though QAT can mitigate this. QAT introduces fake quantization nodes during training, allowing the network to compensate for precision loss. Special training procedures and loss-aware quantization~\cite{choi2018pact} are employed for binary networks. Beyond QAT and PTQ, several quantization schemes have emerged to address different levels of granularity and trade-off. For instance, per-channel quantization adjusts the scale and zero-point for each output channel, leading to better numerical stability and often improved accuracy compared to per-tensor quantization~\cite{banner2019post}. Mixed-precision quantization allows different layers or operations to use different bit-widths (e.g., 8-bit for early layers and 4-bit for later layers), striking a balance between efficiency and performance~\cite{wang2019haq}. Techniques like DoReFa-Net~\cite{zhou2016dorefa} and Learned Step Size Quantization ~\cite{esser2019learned} further improve accuracy by learning optimal quantization parameters during training. In the TinyML context, these methods have enabled the deployment of accurate models like ResNet and MobileNet variants on MCUs with under 256~KB of SRAM. Furthermore, hardware-aware quantization strategies are increasingly integrated into deployment pipelines using tools such as TensorFlow Model Optimization Toolkit \cite{tfmot} and PyTorch's quantization API, enabling automated conversion and validation across platforms.

\subsubsection{Constraints Beyond Quantization} TinyML models often face deployment-specific constraints such as memory budgets (e.g., 64~KB of RAM), latency caps (e.g., 10~ms), and energy limits (e.g., 1~mJ per inference). Frameworks like $\mu$NAS~\cite{chen2021bn} and Once-for-All (OFA)~\cite{cai2019once} support constraint-aware NAS to generate tailored models. In practice, trade-offs are carefully evaluated to balance latency, accuracy, and reliability. Constraint-aware modeling goes beyond architectural choices and encompasses compiler-level and deployment-time optimizations. For example, models must comply with quantization compatibility constraints of hardware accelerators like the GAP8 SoC, which supports only INT8 convolutions~\cite{garofalo2019pulp}, or the Ambiq Apollo3 Blue, which requires careful SRAM and DMA management to maintain sub-mW operation~\cite{ambiqapollo}. Real-time constraints also vary across application domains, such as, voice-triggered devices may tolerate 10--20~ms of latency, whereas anomaly detection in industrial sensors may allow hundreds of milliseconds, but must operate within a strict energy envelope. To handle such variance, modern compilers such as TVM~\cite{chen2018tvm}, Glow \cite{rotem2018glow}, and Apache Relay perform cross-layer optimization and memory layout transformations that respect such deployment constraints. Additionally, tools like MCUNetV2~\cite{lin2021mcunetv2} integrate NAS with firmware-level profiling to co-optimize models for specific MCUs, achieving a better trade-off across the energy-latency-accuracy spectrum.

\subsection{Standard Datasets and Benchmarks}
To enable reproducibility and fair comparison across TinyML approaches, the community relies on a curated set of datasets and benchmarks reflecting the domain's constraints and typical use cases. These benchmarks target tasks such as KWS, low-resolution vision, and anomaly detection.

\thispagestyle{empty}
\begingroup
\fontsize{9pt}{8pt}\selectfont
\begin{table}[h]
  \centering
  \caption{Summary of benchmark datasets and their characteristics}
  \label{tab:benchmark_datasets}
  \renewcommand{\arraystretch}{1.15}
  \begin{tabularx}{\textwidth}{
   >{\raggedright\arraybackslash}p{3.2cm}
  >{\raggedright\arraybackslash}p{2.8cm}
  >{\raggedright\arraybackslash}p{2.3cm}
  >{\raggedright\arraybackslash}p{1.2cm}
  >{\raggedright\arraybackslash}X
  }
    \toprule
    \rowcolor{tinymlblue}
    \textcolor{white}{\textbf{Dataset}} &
    \textcolor{white}{\textbf{Task}} &
    \textcolor{white}{\textbf{Input Size}} &
    \textcolor{white}{\textbf{Classes}} &
    \textcolor{white}{\textbf{Use Case}} \\
    \midrule
    Google Speech Commands~\cite{warden2018speech} & Keyword spotting & 1\,s audio (16\,kHz) & 12--35 & Voice command detection \\
    Visual Wake Words~\cite{chowdhery2019visual}   & Person detection & 96$\times$96 RGB & 2 & Vision-triggered wake-up \\
    Tiny ImageNet~\cite{le2015tiny}                & Image classification & 64$\times$64 RGB & 200 & Low-resolution object recognition \\
    CIFAR-10/100~\cite{krizhevsky2009learning}     & Image classification & 32$\times$32 RGB & 10/100 & Lightweight vision tasks \\
    $\mu$MLPerf~\cite{banbury2020benchmarking}     & Benchmark suite & Various & Various & Standardized TinyML evaluation \\
    \bottomrule
  \end{tabularx}
\end{table}
\endgroup

\subsubsection{Google Speech Commands} Developed by Warden et al., this dataset contains short spoken words sampled at 16~kHz. It is the standard benchmark for KWS. Tasks involve recognizing a fixed vocabulary such as ``\textit{yes}'', ``\textit{no}'', and ``\textit{go}''.

\subsubsection{Visual Wake Words} This dataset is a binary classification task for detecting the presence of a person in low-resolution images. It is used in wake-word-style visual triggers for cameras or embedded vision systems.

\subsubsection{Tiny ImageNet and CIFAR} These datasets serve as benchmarks for image classification under low-resolution and low-memory conditions. Tiny ImageNet is more challenging due to its 200-class design, while CIFAR remains widely used for comparison.

\subsubsection{$\mu$MLPerf Benchmark Suite} MLCommons introduced $\mu$MLPerf to provide standardized evaluation across KWS, image classification, and anomaly detection. It includes metrics like accuracy, model size, memory footprint, and energy per inference, making it one of the most comprehensive benchmarks for TinyML systems.

\section{Challenges and Open Research Problems} 
\label{sec:Challenges}

The field of TinyDL, while demonstrating remarkable progress, faces numerous fundamental challenges that require innovative solutions to realize its full potential across diverse edge computing applications \cite{heydari2025tiny}\cite{adlakha2024challenges}.

\subsection{Trade-off Between Accuracy and Footprint}

The fundamental tension between model accuracy and resource consumption represents the most persistent challenge in TinyDL deployment \cite{svoboda2022deep}. Current approaches often sacrifice significant accuracy to meet stringent memory and computational constraints, limiting the applicability of TinyDL systems in accuracy-critical applications \cite{svoboda2022deep}.

\subsubsection{Memory Hierarchy Complexity}

The complex memory hierarchy of MCUs, with limited SRAM (typically 320KB) and flash storage (1MB), creates intricate optimization challenges \cite{svoboda2022deep}. Traditional ML benchmarks assume gigabytes of memory, making direct adaptation impossible \cite{svoboda2022deep}. The memory bottleneck affects not only model storage but also intermediate activations during inference, requiring sophisticated memory scheduling strategies that consider the entire network topology rather than layer-wise optimization \cite{svoboda2022deep}.

\subsection{Computational Efficiency Paradox}

While larger models generally achieve better accuracy, the computational resources available on MCUs create hard limits on deployable model complexity \cite{svoboda2022deep}. The emergence of transformer architectures exacerbates this challenge, as the quadratic complexity of self-attention mechanisms conflicts with the linear resource scaling capabilities of edge devices \cite{jung2024optimizing}. Current solutions like attention approximation and sparse attention patterns show promise but remain insufficient for complex real-world applications \cite{jung2024optimizing}.

\subsection{Secure Model Updates in the Field}

The deployment of TinyDL models in remote and potentially hostile environments creates unprecedented security challenges that traditional cloud-based ML systems do not encounter \cite{shah2024enhancing}\cite{huckelberry2024tinyml}.

\subsubsection{Adversarial Attack Transferability}

Research demonstrates that adversarial attacks crafted on powerful host machines can successfully transfer to resource-constrained devices like ESP32 and Raspberry Pi, highlighting the vulnerability of TinyML systems to security threats \cite{shah2024enhancing}. The limited defensive capabilities of edge devices make them particularly susceptible to model extraction and evasion attacks, where adversaries can potentially reconstruct sensitive model parameters or manipulate model behavior \cite{shah2024enhancing}\cite{huckelberry2024tinyml}.

\subsubsection{Model Update Vulnerabilities}

TinyML devices often operate in physically accessible environments where attackers can potentially intercept or manipulate model updates \cite{huckelberry2024tinyml}. The limited computational resources of these devices make it challenging to implement robust cryptographic protocols for secure model transmission and verification \cite{huckelberry2024tinyml}. The integration of hardware security modules and trusted execution environments into TinyML platforms represents a promising direction, but current solutions significantly increase cost and power consumption \cite{pujari2023enhancing}.

\subsection{Generalization Under Few-Shot Learning}

The limited computational and storage resources of TinyML devices severely constrain their ability to adapt to new tasks or domains through traditional machine learning approaches \cite{parnami2022learning}\cite{kim2025few}. For example, wearable health-monitoring devices deployed in elderly care settings often need to adapt to new users with limited labeled data. Supporting such personalized learning on-device without cloud retraining demands efficient few-shot learning techniques that operate within sub-256~KB memory and minimal latency budgets.

\subsubsection{Meta-Learning Constraints}

While few-shot learning techniques show promise for enabling rapid adaptation in resource-constrained environments, the meta-learning algorithms themselves often require significant computational resources that exceed TinyML capabilities \cite{parnami2022learning}. The memory requirements for maintaining meta-parameters and adaptation mechanisms conflict with the storage limitations of MCUs, necessitating novel approaches to meta-learning specifically designed for edge deployment \cite{kim2025few}.

\subsubsection{Continual Learning Limitations}

The ability to learn continuously from new data while retaining previously acquired knowledge represents a critical capability for long-term TinyML deployment \cite{fini2020online}. However, the memory limitations of edge devices make it challenging to implement effective continual learning strategies that prevent catastrophic forgetting while enabling knowledge acquisition from limited data samples \cite{fini2020online}. Recent research on memory-constrained online continual learning demonstrates that effective continual learning is possible under severe memory limitations, but requires algorithmic innovations that fundamentally differ from traditional approaches \cite{fini2020online}.

\subsection{Lack of Standardized Benchmarks for TinyDL}

The absence of comprehensive, standardized benchmarks specifically designed for TinyDL systems impedes systematic progress and fair comparison of different approaches \cite{banbury2021mlperf}.

\subsubsection{Hardware Heterogeneity Challenges}

The diverse landscape of TinyML hardware platforms, ranging from ARM Cortex-M MCUs to specialized AI accelerators, complicates the development of universally applicable benchmarks \cite{banbury2021mlperf}\cite{chang2023benchmarking}. Each platform exhibits unique characteristics in terms of memory hierarchy, instruction sets, and optimization opportunities, making it difficult to establish fair comparison metrics across different systems \cite{banbury2021mlperf}.

\subsubsection{Multi-Objective Evaluation Complexity}

Traditional ML benchmarks focus primarily on accuracy metrics, but TinyDL systems require evaluation across multiple dimensions including latency, energy consumption, memory usage, and model size \cite{banbury2021mlperf}. The MLPerf Tiny benchmark provides valuable baseline comparisons but focuses primarily on computer vision and simple audio processing tasks, neglecting other important application domains such as natural language processing and sensor fusion \cite{banbury2021mlperf}.

\subsection{Limited Tool Support for Advanced DL on MCUs}

The deployment of sophisticated deep learning models on MCUs faces significant toolchain limitations that restrict the practical implementation of state-of-the-art architectures \cite{svoboda2022deep}.

\subsubsection{Transformer Deployment Challenges}
While transformer architectures represent the state-of-the-art in many AI applications, their deployment on MCUs remains severely limited by inadequate compiler and runtime support \cite{jung2024optimizing}. Current toolchains lack optimized implementations of attention mechanisms and layer normalization operations, forcing researchers to develop custom, platform-specific solutions that limit portability and reproducibility \cite{jung2024optimizing}. This limitation is particularly evident in tasks such as on-device natural language understanding or voice assistant applications, where lightweight models like TinyBERT or DistilBERT cannot be fully deployed on common MCUs (e.g., ARM Cortex-M4) due to the absence of efficient attention-layer primitives in existing toolchains, such as TensorFlow Lite Micro or CMSIS-NN.

\subsubsection{Cross-Platform Portability}

The heterogeneous nature of TinyML hardware creates significant challenges for developing portable software solutions \cite{svoboda2022deep}. Current toolchains often require platform-specific optimizations that limit code reusability and increase development overhead, hindering the widespread adoption of TinyDL techniques across diverse hardware platforms \cite{svoboda2022deep}.

\bigskip

These challenges represent fundamental research opportunities that will shape the future development of TinyDL\cite{heydari2025tiny}\cite{gill2025edge}. Addressing these issues requires interdisciplinary collaboration across machine learning, computer systems, and hardware design communities to develop innovative solutions that unlock the full potential of edge AI applications \cite{gill2025edge}\cite{wang2025empowering}.

\section{Future Directions} 
\label{sec:Future}

As TinyDL systems mature and expand across diverse application domains from healthcare and smart homes to industrial automation and autonomous sensing, new challenges and technological frontiers are emerging. Addressing these will require interdisciplinary advances in hardware design, algorithmic efficiency, secure training, and adaptable software ecosystems. This section outlines five promising directions for future exploration. Neuromorphic architectures employing Spiking Neural Networks (SNNs) offer an alternative computational paradigm designed for ultra-low-power, event driven processing typical of brain-inspired systems. These architectures promise efficient always-on inference on MCU-scale devices ideal for continuous monitoring applications-with hardware platforms like BrainChip’s Akida and Intel’s Loihi leading the way. To fully leverage SNNs in TinyDL, research must advance surrogate-gradient training, event encoding techniques, and software toolchains that map spiking models onto neuromorphic hardware seamlessly \cite{liang2021pruning}.

Implementing federated learning (FL) in TinyDL contexts addresses privacy and adaptability by enabling decentralized learning across devices without sharing raw data crucial for distributed sensor networks. Lightweight frameworks such as TinyFedTL and TinyMetaFed demonstrate on-device aggregation of quantized updates, yet challenges remain in managing communication overhead, heterogeneous device capabilities, and adversarial resilience \cite{zhangtraining, huckelberry2024tinyml}. Future work must focus on sparsified updates, asynchronous or hierarchical FL protocols, and secure aggregation mechanisms amenable to TinyML constraints.

Tiny Foundation Models refer to miniaturized versions of large pretrained models intended for deployment on edge hardware. Promising techniques such as knowledge distillation, structured pruning, and quantizationapplied to models like TinyViT have shown the potential to reduce model size to MCU-suitable scales while preserving task performance \cite{banbury2021micronets, lin2020mcunet}. The next step is to enable modular foundational architectures, where a general “backbone” pre-trained model supports multiple lightweight task-specific heads, with workflows powered by on-device or Edge AutoML-enabled fine-tuning. Edge AutoML seeks to automate the process of designing, compressing, and deploying TinyDL models on resource-constrained devices. Techniques like hardware-aware NAS frameworks, such as TinyNAS and Once-for-All Networks, have demonstrated effective ways to balance accuracy with memory and latency constraints \cite{shafiee2017squishednets, jung2024optimizing}. However, integrating AutoML into full deployment pipelines remains an open challenge. Future research should focus on combining AutoML with model compression strategies like quantization and pruning and incorporating hardware feedback to generate models that are not only accurate but also energy-efficient and deployable in real-world TinyDL scenarios.

Domain-specific accelerators, including NPUs, ASICs, FPGAs, and specialized RISC-V engines, offer substantial gains in inference speed, energy efficiency, and model scalability for TinyDL. Devices like the EdgeTPU and transformer-focused RISC-V extensions efficiently deliver quantized convolution and attention workloads, outperforming general-purpose MCUs \cite{jung2024optimizing, liang2021pruning}. The challenge now is to develop advanced compilation toolchains that partition and schedule TinyDL models across heterogeneous hardware, integrate with platforms like TFLite Micro and CMSIS-NN, and maximize runtime configuration flexibility without sacrificing portability or ease of development \cite{Liu2023MicroTVM, rotem2018glow}.


\section{Conclusions}
\label{sec:conclusion}

This survey presents a comprehensive examination of the evolution from TinyML to TinyDL, highlighting how the convergence of efficient model architectures, software toolchains, and hardware platforms has enabled sophisticated on-device intelligence in severely resource-constrained environments. We begin by delineating the scope and distinction between TinyML and TinyDL, emphasizing the growing need to embed deep learning capabilities, once reserved for data centers, into low-power MCUs and edge devices. We have outlined the hardware advancements, including the emergence of neural accelerators and specialized ASICs, that now support the deployment of deep networks with kilobyte-scale memory footprints and milliwatt power budgets. Simultaneously, we explored the critical role of model optimization techniques such as quantization, pruning, and joint compression strategies, as well as the contributions of NAS in tailoring architectures to edge constraints. On the software side, we cataloged an extensive range of deployment frameworks, compiler toolchains, and AutoML platforms that streamline the end-to-end TinyDL lifecycle. Through domain-specific applications in vision, audio, healthcare, and industrial monitoring, we demonstrated the transformative potential of TinyDL across sectors demanding low latency, energy efficiency, and data privacy.

Looking ahead, TinyDL is poised to catalyze a new generation of edge-native intelligence. This includes the development of neuromorphic architectures using spiking neural networks, federated learning for decentralized personalization, and ultra-lightweight foundation models capable of generalization across tasks and modalities. The co-design of hardware and software will become increasingly central, as will the creation of standardized, energy-aware benchmarks to evaluate system performance holistically. By bridging the conceptual, architectural, and practical aspects of TinyDL, this survey aims to serve as a foundational resource for both researchers and practitioners. It underscores the critical shift from cloud dependence to autonomous, efficient edge intelligence, laying the groundwork for continued innovation in AI at the very edge of computing.




\appendix

\end{document}